\documentclass{article}

\usepackage[preprint]{neurips_2026}

\usepackage[utf8]{inputenc}
\usepackage[T1]{fontenc}
\usepackage{hyperref}
\hypersetup{
  pdftitle={PinPoint: Training-Free Referring Segmentation via Prompt Disambiguation},
  pdfauthor={Pouya Sadeghi, Shawn He, Pablo Guerrero, C Thomas, Alex Wong, Sirisha Rambhatla},
  pdfkeywords={referring image segmentation, vision-language models, SAM, training-free, prompt disambiguation}
}
\usepackage{url}
\usepackage{booktabs}
\usepackage{amsfonts}
\usepackage{amsmath}
\usepackage{nicefrac}
\usepackage{microtype}
\usepackage{xcolor}
\usepackage{graphicx}
\usepackage{float}
\usepackage{capt-of}
\usepackage{subcaption}
\usepackage{multirow}
\usepackage{enumitem}
\usepackage[most]{tcolorbox}
\usepackage{algorithm}
\usepackage{algpseudocode}

\graphicspath{{figures/}}

\newcommand{\methodshort}{PinPoint}
\newcommand{\methodlong}{Prompting with Informative Interior Points}
\newcommand{\method}{\textbf{\methodshort}}

\newcommand{\best}[1]{\textbf{#1}}
\newcommand{\sbest}[1]{\underline{#1}}

\definecolor{posgreen}{HTML}{1B7837}
\definecolor{negred}{HTML}{B2182B}
\newcommand{\dpos}[1]{\textcolor{posgreen}{#1}}
\newcommand{\dneg}[1]{\textcolor{negred}{#1}}

\title{\methodshort{}: Training-Free Referring Segmentation\\via Prompt Disambiguation}

\author{%
\begin{tabular}{c}
Pouya Sadeghi\textsuperscript{1,2} \quad
Shawn He\textsuperscript{3} \quad
Pedro Pablo Guerrero Vela\textsuperscript{3} \quad
C. Thomas\textsuperscript{3} \\
Alex Wong\textsuperscript{3} \quad
Sirisha Rambhatla\textsuperscript{1,2}
\end{tabular}
\\[1em]
\begin{tabular}{c}
\textsuperscript{1}University of Waterloo \qquad
\textsuperscript{2}Critical ML, University of Waterloo \qquad
\textsuperscript{3}Apple
\end{tabular}
\\[0.4em]
\begin{tabular}{c}
\texttt{\{pouya.sadeghi,sirisha.rambhatla\}@uwaterloo.ca}
\end{tabular}
}

\begin{document}

\maketitle

\begin{abstract}
Modern referring image segmentation pipelines couple a vision-language model (VLM) for grounding with a promptable segmenter such as the Segment Anything Model (SAM) for mask generation.
Prior training-free instances of this recipe consistently trail fine-tuned and reinforcement-learning (RL)-tuned specialists, and it has been unclear whether the gap comes from the VLM's grounding, SAM's capacity, or the prompt. We show that the gap is dominated by \emph{prompt ambiguity}: a VLM-proposed bounding box (bbox) leaves SAM to guess which pixels inside the bbox belong to the object the expression denotes.
Interior points are the natural disambiguator, but where they fall matters; prior work relies on naively sampled points that land on boundaries, distractors, and background clutter, and can even \emph{hurt} performance compared to the bbox alone.
Supervised and RL-tuned methods close this gap by training a VLM to predict better points; we show that this training is unnecessary.
At a matched budget of five interior points, replacing naive sampling with stable, informative point selection improves cumulative Intersection-over-Union (cIoU) by 12–18 points across RefCOCO/+/g, with every model fixed.
We turn this observation into \method{}, a deterministic, training-free point selector that fuses four visual cues into a consensus map, selects compact, spatially diverse points away from boundaries, and uses the frozen VLM to label each point.
Without any task-specific training, \methodshort{} matches supervised and RL-tuned specialists on the same stack while issuing only two VLM calls per query.

\end{abstract}


\section{Introduction}
\label{sec:introduction}

\begin{figure}[t]
    \centering
    \includegraphics[width=0.99\textwidth]{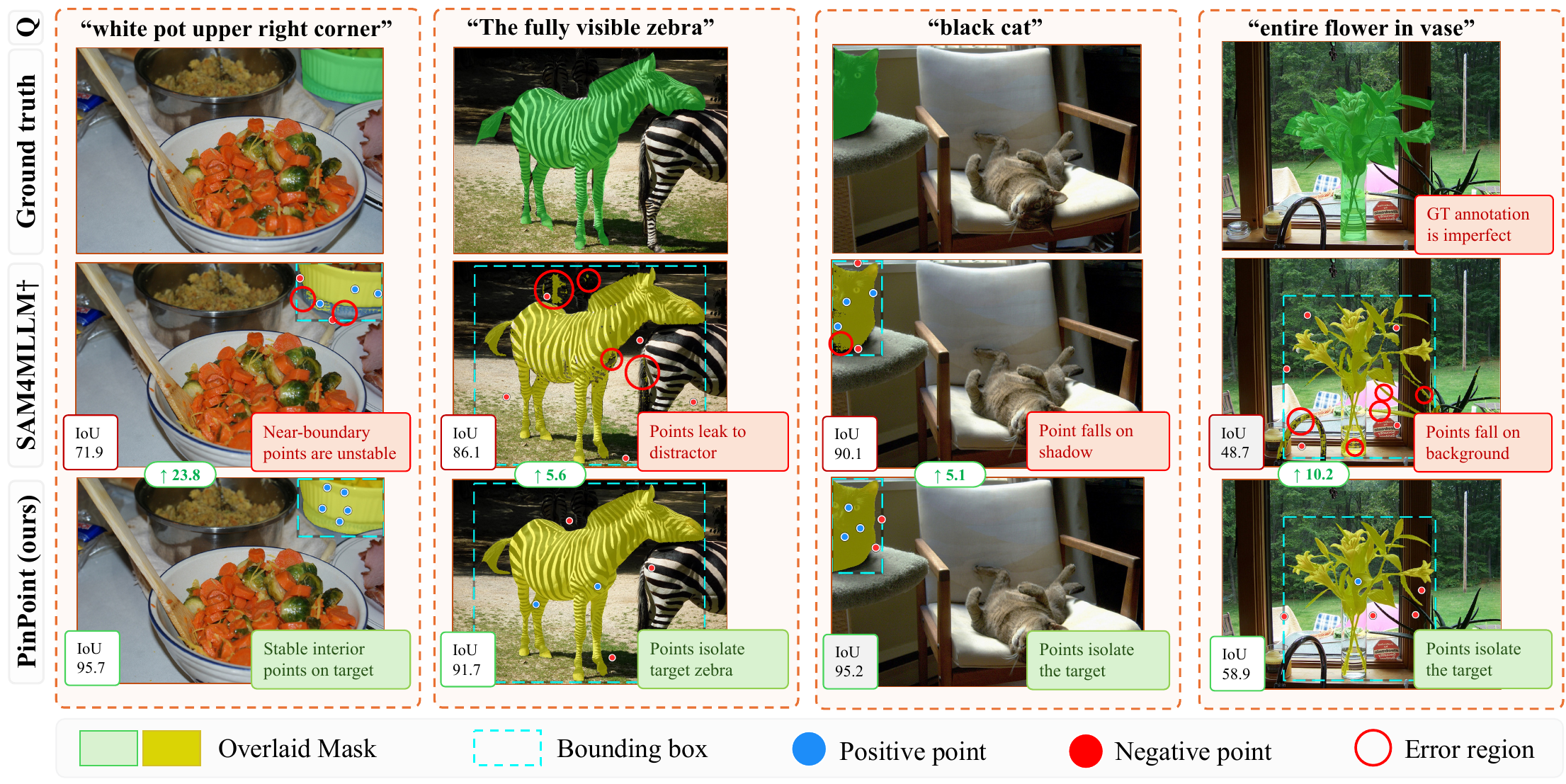}
    \caption{\textbf{Prompt disambiguation drives the training-free referring segmentation gap.} For each query (columns), we show the ground truth (top), matched-budget SAM4MLLM$^\dagger$ (middle), and \method{} (bottom). All methods use the same frozen VLM, point budget ($N{=}5$), and single SAM forward pass; only the point selector changes. IoU is per-example mask overlap (higher is better). $\dagger$ denotes our training-free reproduction of SAM4MLLM's PQPP setting: uniformly sampled interior points (naive sampling) labeled by the frozen VLM.}
    \label{fig:teaser}
\end{figure}

Referring image segmentation generates a pixel-level binary mask of the object that a free-form natural-language expression denotes~\citep{refcoco2014emnlp, refcocoplus2017cvpr, refcocog2016cvpr}. Unlike traditional semantic or instance segmentation, which depend on a predefined label set~\citep{MaskRCNN2017ICCV, Mask2Former2022CVPR, MultimodalRISSurvey2025}, referring segmentation must handle compositional, attribute-rich expressions such as \textit{``the partially hidden zebra''} or \textit{``the white pot in the upper right corner''}. This setting better reflects how humans describe objects in real-world use cases such as instruction-following agents~\citep{CogAgent2024CVPR} and image-editing tools~\citep{InstructPix2Pix2023CVPR, InpaintAnything2023}, while imposing higher demands on the model: nuanced grounding of language to image regions, fine-grained visual perception, and precise pixel-level mask generation. Modern systems for this task have converged on a modular recipe that pairs two pretrained foundation models: a vision-language model (VLM)~\citep{LLaVA2023NeurIPS, Qwen25VL2025, Qwen3VL2025} for semantic grounding, and a promptable segmenter such as the Segment Anything Model (SAM)~\citep{SAM2023ICCV, SAM2_2024} for pixel-level mask generation~\citep{LISA_2024_CVPR, sam4mllm2024eccv, GSVA2024CVPR, SegZero2025arxiv, SAMR1_2025, SAMVeteran2025}.

This recipe is appealing because it reuses two strong off-the-shelf foundation models. However, the interface between them is fragile: prior training-free instances consistently remain behind their fine-tuned and RL-tuned counterparts on benchmarks by a noticeable margin and it has not been clear whether the gap lies in the VLM’s grounding, in SAM’s capacity, or in the prompt that connects them. We argue that the bottleneck is the prompt itself, a phenomenon we call \emph{prompt ambiguity}: a VLM-predicted bounding box is an under-specified region that contains the target alongside partial occluders, distractors, boundaries, and background clutter~\citep{GRES2023CVPR}. This leaves SAM to guess which pixels inside it belong to the referent. SAM, in turn, is the least stable near object boundaries~\citep{quantifyinglimitssegmentationfoundation2024arxiv, StableSAM2025ICLR, ProSAM2025ICCV}, and its mask predictions degrade when prompted with redundant or off-target points~\citep{StableSAM2025ICLR, SAMAug2024Arxiv}; this makes \emph{the choice of interior points} a critical question. Figure~\ref{fig:teaser} makes this visible: at a matched point budget, na\"{i}ve interior sampling places points on boundaries and nearby distractors, while \methodshort{} selects stable points that isolate the target.

Existing work closes this gap through additional training, falling into three broad categories. The first adds segmentation-aware tokens to the VLM so that it can emit mask-shaped outputs~\citep{LISA_2024_CVPR, PixelLM2024CVPR, VISA2024ECCV, GSVA2024CVPR, PerceptionGPT2024CVPR, Sa2VA2025, UniPixel2025NeurIPS}; auxiliary threads layer chain-of-thought, decomposition, or mask-feedback loops on top~\citep{thinkFirst2025Arxiv, resanything2025neurips, SegLLM2024arxiv, CoT-RVS2025arxiv}. The second fine-tunes the VLM to produce SAM-compatible point prompts directly~\citep{sam4mllm2024eccv}. The third replaces supervision with reinforcement learning, post-training a VLM with GRPO on a few thousand curated samples to emit better single-shot points or refine them iteratively~\citep{SegZero2025arxiv, SAMR1_2025, SAMVeteran2025, LENS2026AAAI}. Across all three families, the recurring object of optimization is the same: produce a better spatial prompt for SAM. A simpler question has not been answered yet: do these informative points actually require learning, or can they be proposed without any training?

We argue that the points do not need to be \emph{generated} by a trained policy; they can be \emph{selected} from image-side evidence and \emph{verified} by the frozen VLM. Classical visual cues (saliency, local edges, entropy, and a spatial prior) already encode where reliable evidence sits inside a coarse box, and a frozen VLM is a competent per-point verifier. Prompt disambiguation, in this view, is not a learning problem but a deterministic, image-side computation.

We turn this idea into \method{} (\emph{\methodlong}), a deterministic, training-free point selector that converts a coarse VLM bounding box into a clarifying composite prompt: the bounding box plus a small set of semantically labeled interior points. Inside each predicted box, \methodshort{} fuses visual cues into a consensus map, extracts a compact and spatially diverse interior point set, and reuses the frozen VLM to label each point as positive or negative; SAM is then run once with the resulting prompt, and the entire pipeline costs only two VLM calls per query. At a matched point budget across RefCOCO/+/g, swapping naive sampling for \methodshort{} improves cIoU by 12--18 points with every model frozen (Table~\ref{tab:main_tf}); on the same stack used by recent SFT and RL specialists, \methodshort{} is competitive with all of them without a single task-specific gradient update (Section~\ref{sec:experiments}). Together, these results identify \emph{prompt ambiguity} as the dominant remaining bottleneck in modular VLM+SAM referring segmentation, and show that the post-training step recent methods perform to learn point selection is replaceable by a principled image-side computation.


\section{Related Work}
\label{sec:related_work}

\textbf{Supervised fine-tuning.}
A large body of work equips VLMs with segmentation-specific output interfaces, typically by adding mask-producing heads or segmentation tokens that decode into pixel masks~\citep{LISA_2024_CVPR, PixelLM2024CVPR, VISA2024ECCV, GSVA2024CVPR, PerceptionGPT2024CVPR, Sa2VA2025, UniPixel2025NeurIPS}, with reasoning and feedback-loop variants layering chain-of-thought, decomposition, or mask-feedback on top~\citep{thinkFirst2025Arxiv, resanything2025neurips, SegLLM2024arxiv, CoT-RVS2025arxiv}. These methods change the \emph{output head} of the VLM, not the prompt fed to SAM. The closest training-based work to ours is \citet{sam4mllm2024eccv}: it uniformly samples candidate interior points, and supervises the VLM to ground the query and label each extracted point. Their results show that VLM-as-\emph{labeler} outperforms VLM-as-\emph{generator}, which suggests \emph{select-then-label} as the regime worth pursuing. \methodshort{} reuses select-then-label \emph{without the supervised tuning step}.

\textbf{Reinforcement learning for prompt policies.}
A more recent line keeps the frozen-segmentor design and swaps supervision for reinforcement learning, shaping the VLM's prompt with rewards derived from the box or mask. Single-shot variants train the VLM to emit better bounding boxes and points in one pass~\citep{SegZero2025arxiv, LENS2026AAAI}, with \citet{SAMR1_2025} deriving the reward directly from the resulting SAM mask. \citet{SAMVeteran2025} go further, casting the VLM as an iterative agent that proposes refinement points and learns when to terminate the refinement loop. The recurring object of optimization across all variants is a learned policy in the language model that emits or refines the spatial prompt. \methodshort{} \emph{removes the learned policy entirely} by selecting stable and informative points off the image with a deterministic procedure and reusing the frozen VLM to label them.

\textbf{Training-free prompt disambiguation.}
\citet{GRES2023CVPR} first frames the bounding box as an ambiguous spatial signal, and a growing empirical body characterizes when SAM's mask becomes unstable: it is least reliable near object boundaries, degrades under redundant or off-target prompts, and benefits from explicit uncertainty modeling~\citep{quantifyinglimitssegmentationfoundation2024arxiv, StableSAM2025ICLR, ProSAM2025ICCV}. Existing responses to this evidence either operate on the input side without selecting better points to begin with~\citep{SAMAug2024Arxiv}, or sidestep promptable segmentation altogether by refining CLIP Grad-CAM heatmaps outside the VLM+SAM regime~\citep{IteRPrimE2025AAAI}. Prior training-free work thus diagnoses prompt sensitivity or escapes it by changing the regime; \methodshort{} \emph{acts on the same evidence from inside the select-then-label pipeline}, with a deterministic, image-side selector replacing learned point selection and the frozen VLM to verify each candidate.


\section{Method}
\label{sec:method}

\begin{figure}[t]
    \centering
    \includegraphics[width=0.99\textwidth]{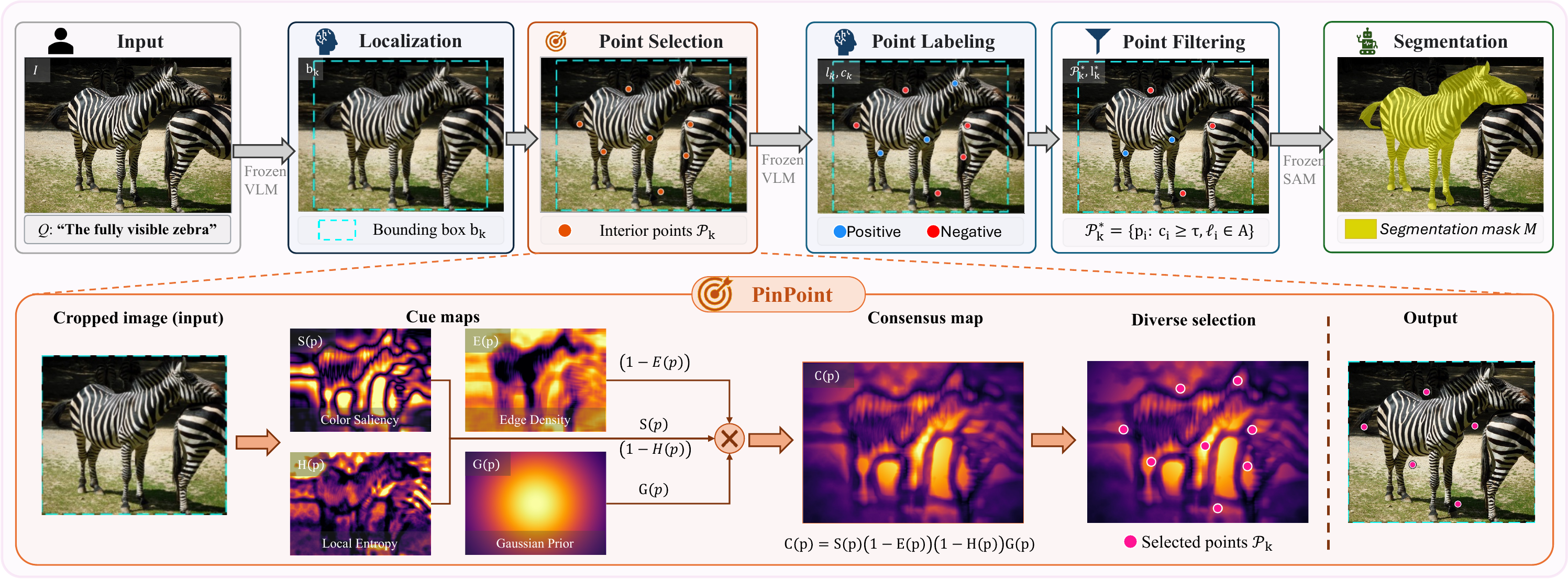}
    \caption{\textbf{Training-free referring segmentation with \methodshort{}.} A frozen VLM localizes the referent with candidate bboxes; \methodshort{} selects informative interior points inside each bbox; the same frozen VLM labels each point as positive or negative; confidence filtering removes unreliable labels; SAM produces the final mask in a single forward pass. The lower inset shows the \methodshort{} selector: cue maps computed on a crop, fused into a consensus map, and converted into a diverse interior point set.}
    \label{fig:pipeline}
\end{figure}

\methodshort{} disambiguates the prompt passed to SAM \emph{before} mask generation. Each VLM-predicted bounding box is converted into a composite prompt of \emph{bounding box (bbox) + a small set of semantically labeled interior points}; with all backbones frozen and the point budget matched, the choice of these interior points has a large effect on cIoU (Section~\ref{sec:experiments}).

\subsection{Problem setup}
\label{sec:method:setup}

We are given an image $I \in \mathbb{R}^{H \times W \times 3}$ and a natural-language query $Q$ that may specify zero, one, or multiple visible target instances. The goal is a binary mask $M \in \{0,1\}^{H \times W}$ whose positive pixels correspond to the object(s) denoted by $Q$ (or the all-zero mask when there is no visible referent). We construct $M$ through a five-stage pipeline (Figure~\ref{fig:pipeline}):

\begin{equation}
    (I, Q) \xrightarrow{\text{Localize}} \mathcal{B} \xrightarrow{\text{\methodshort}} \{(b_k, \mathcal{P}_k)\} \xrightarrow{\text{Label}} \{(b_k, \mathcal{P}_k, \boldsymbol{\ell}_k, \mathbf{c}_k)\} \xrightarrow{\text{Filter}} \{(b_k, \mathcal{P}_k^{\star}, \boldsymbol{\ell}_k^{\star})\} \xrightarrow{\text{SAM}} M.
\label{eq:pipeline}
\end{equation}

\emph{Localization} turns $(I,Q)$ into candidate bboxes $\mathcal{B} = \{b_1,\dots,b_K\}$. \methodshort{} extracts a small set of interior points $\mathcal{P}_k$ inside each $b_k$. \emph{Point labeling} attaches a label $\ell_{k,i} \in \{\texttt{positive},\texttt{negative}\}$ and confidence $c_{k,i} \in [0,1]$ to every $p_{k,i} \in \mathcal{P}_k$ via the same VLM; we collect these into the per-bbox tuples $\boldsymbol{\ell}_k = (\ell_{k,i})_{p_{k,i} \in \mathcal{P}_k}$ and $\mathbf{c}_k = (c_{k,i})_{p_{k,i} \in \mathcal{P}_k}$. \emph{Filtering} keeps the subset $\mathcal{P}_k^{\star} \subseteq \mathcal{P}_k$ that passes a confidence threshold and a label allowlist, with $\boldsymbol{\ell}_k^{\star}$ the corresponding labels restricted to $\mathcal{P}_k^{\star}$. \emph{Segmentation} hands $(b_k, \mathcal{P}_k^{\star}, \boldsymbol{\ell}_k^{\star})$ to SAM in a single forward pass, and the per-bbox masks are combined into $M$.

\subsection{Localization}
\label{sec:method:localization}
The first stage uses the frozen VLM as a grounded proposer that turns $(I,Q)$ into a set of bboxes $\mathcal{B} = \{b_1,\dots,b_K\}$ in pixel coordinates, where $K$ may be $0$ (no detected referent), $1$, or larger. We use grammar-guided decoding~\citep{vllm2023sosp} to constrain the VLM's output to valid JSON to avoid fragile text post-processing; the prompt and EBNF grammar are in Appendix~\ref{app:prompts_localization}. We only require this stage to provide \emph{coarse} spatial support; a predicted bbox may include background clutter or overlap nearby instances, and therefore remain an under-specified prompt for SAM.

\subsection{Point selection via \methodshort{}}
\label{sec:method:point_selection}

\methodshort{} is applied inside each predicted bbox to convert this coarse signal into an informative prompt. A useful point prompt should identify stable visual evidence inside the localized region while avoiding regions where SAM and the VLM verifier are unreliable; SAM is least stable near object boundaries~\citep{quantifyinglimitssegmentationfoundation2024arxiv} and its mask predictions degrade when prompted with redundant or off-target points~\citep{StableSAM2025ICLR, SAMAug2024Arxiv}. Given a bbox $b$ and a budget $N$, \methodshort{} computes four pixel-wise cues over an extended crop, fuses them into a consensus score, and greedily selects up to $N$ spatially diverse points inside $b$.

\paragraph{Cues.} We score each pixel using four complementary cues (full formulas in Appendix~\ref{app:cue_definitions}): \emph{color-contrast saliency} $S$, the $\ell_2$ distance between mean CIELAB colors in a local disk and an annulus around $p$~\citep{ittikoch1998}; \emph{edge density} $E$, a box-averaged Sobel gradient magnitude flagging boundary-adjacent pixels; \emph{local Shannon entropy} $H$ over quantized gradient magnitudes, identifying textural clutter; and a \emph{Gaussian spatial prior} $G$ centered at the bbox centroid with bandwidth proportional to the bbox diagonal. Cues are computed over an extended crop $b_{\text{ext}}$ obtained by extending $b$ by a factor $\alpha$. Each raw map is rescaled to $[0,1]$ via a percentile-clipped min--max normalizer $\mathcal{N}(\cdot)$ (Eq.~\ref{eq:norm} in the appendix); we reuse $S,E,H,G$ for the normalized cues. The cues are classical signals from the saliency literature~\citep{harelGBVS2006, hou2007spectralresidual, juddcenterbias2009}; \methodshort{}'s contribution is how they are composed: inside each VLM-proposed bbox, coupled to a downstream verifier, with a budget tied to cue agreement through Soft-NMS~\citep{soft-nms2017ICCV} so the selector can return fewer than $N$ points when reliable evidence is sparse.

\paragraph{Fusion.} We combine the normalized cues into a consensus score map $C$ using one of two fusion modes. The \emph{multiplicative consensus} variant requires all cues to agree:
\begin{equation}
    C_{\text{mult}}
    \;=\;
    S \,\times\, (1-E) \,\times\, (1-H) \,\times\, G,
    \label{eq:consensus_mult}
\end{equation}
so that a pixel scores high only if it is visually distinctive, away from strong boundaries, locally homogeneous, and spatially plausible; a weak value in any cue suppresses the score, making this variant conservative in cluttered scenes. The \emph{max-saliency-smooth} variant keeps saliency as the dominant signal and uses edge density and entropy as smoothness guards:
\begin{equation}
    C_{\text{msm}} \;=\; S \;\times\; \max\!\big(\,1 - \lambda_E E - \lambda_H H,\ 0\,\big) \;\times\; G,
    \label{eq:consensus_msm}
\end{equation}
where $\lambda_E,\lambda_H \geq 0$ control the strength of the edge and entropy penalties; textured but salient interior regions can still receive high scores while pixels with strong boundary evidence or heavy clutter are zeroed out by the $\max(\cdot,0)$ rectifier. Both variants share the factored form $C = (\text{visual evidence}) \times G$ but differ in how strongly the smoothness cues gate the saliency signal: $C_{\text{mult}}$ enforces strict conjunction, while $C_{\text{msm}}$ shows that the strict-consensus constraint can be loosened without loss.

\paragraph{Candidate extraction and diversity selection.} We extract local maxima of $C$ as the \emph{candidate pool} $\widetilde{\mathcal{P}} \subset \operatorname{interior}(b)$ with initial scores $C(p)$ (cue maps live on the extended crop, but candidates are restricted to the original bbox), and grow a \emph{selected set} $\mathcal{P}$ greedily from $\widetilde{\mathcal{P}}$. To avoid redundant prompts, the next point is chosen by a Soft-NMS-style~\citep{soft-nms2017ICCV} virtual score $\widetilde{C}_{\mathcal{P}}(p)$ that multiplies $C(p)$ by a Gaussian penalty in distance (with bandwidth $\sigma_{\text{nms}}$) to the nearest point already in $\mathcal{P}$ (Eq.~\ref{eq:soft_nms} in Appendix~\ref{app:cue_definitions}). The first selection reduces to $\arg\max_p C(p)$; after each selection, virtual scores are updated, so nearby candidates are suppressed while distant high-scoring candidates remain competitive. The procedure stops in three cases: $N$ points have been selected, no candidates remain, or every remaining virtual score has fallen below the floor $\varepsilon$. \methodshort{} can therefore return fewer than $N$ points rather than padding with weak or redundant ones. \methodshort{} adds no VLM calls beyond localization and labeling, and its CPU cost is dominated by the surrounding VLM and SAM forward passes (Section~\ref{sec:experiments:efficiency}).

\begin{algorithm}[t]
\caption{\methodshort{} point selection inside a single bounding box. Defaults for all hyperparameters are listed in Appendix~\ref{app:reproducibility}.}
\label{alg:phips}
\begin{algorithmic}[1]
\State \textbf{Input:} image $I$, bounding box $b$, budget $N$, fusion mode $m \in \{\text{mult}, \text{msm}\}$
\State \textbf{Output:} selected points $\mathcal{P} \subset \operatorname{interior}(b)$ with $|\mathcal{P}| \leq N$
\State $b_{\text{ext}} \gets \operatorname{Extend}(b, \alpha)$
\For{each cue $X \in \{S, E, H, G\}$}
    \State compute raw cue map $X_{\text{raw}}$ over $b_{\text{ext}}$ \Comment{Eqs.~\ref{eq:cue_S}--\ref{eq:cue_G}}
    \State $X \gets \mathcal{N}(X_{\text{raw}})$ \Comment{Eq.~\ref{eq:norm}}
\EndFor
\State $C \gets \operatorname{Fuse}(S, E, H, G, m)$ \Comment{Eq.~\ref{eq:consensus_mult} or Eq.~\ref{eq:consensus_msm}}
\State $\widetilde{\mathcal{P}} \gets \operatorname{LocalMaxima}(C) \cap \operatorname{interior}(b)$
\State $\mathcal{P} \gets \emptyset$
\While{$|\mathcal{P}| < N$ \textbf{and} $\widetilde{\mathcal{P}} \setminus \mathcal{P} \neq \emptyset$}
    \State compute $\widetilde{C}_{\mathcal{P}}(p)$ for all $p \in \widetilde{\mathcal{P}} \setminus \mathcal{P}$ \Comment{Eq.~\ref{eq:soft_nms}; $\widetilde{C}_{\emptyset}(p)=C(p)$}
    \If{$\max_{p \in \widetilde{\mathcal{P}} \setminus \mathcal{P}} \widetilde{C}_{\mathcal{P}}(p) \leq \varepsilon$}
        \State \textbf{break}
    \EndIf
    \State $p^\star \gets \operatorname*{arg\,max}_{p \in \widetilde{\mathcal{P}} \setminus \mathcal{P}} \widetilde{C}_{\mathcal{P}}(p)$
    \State $\mathcal{P} \gets \mathcal{P} \cup \{p^\star\}$
\EndWhile
\State \Return $\mathcal{P}$
\end{algorithmic}
\end{algorithm}

\subsection{Point labeling}
\label{sec:method:point_labeling}

\methodshort{}'s points are geometrically interior but carry no semantic understanding of the target. We use the same frozen VLM as a per-point \emph{semantic verifier}: for each candidate $p_{k,i} \in \mathcal{P}_k$, the VLM receives $(I, Q, b_k, p_{k,i})$ and returns a label and a verbalized confidence,
\begin{equation}
    (\ell_{k,i},\, c_{k,i}) \;=\; g_{\text{ver}}\!\bigl(I,\, Q,\, b_k,\, p_{k,i}\bigr),
    \qquad \ell_{k,i}\in\{\texttt{positive},\texttt{negative}\},\ c_{k,i} \in [0,1],
    \label{eq:verifier}
\end{equation}
where $\ell_{k,i}=\texttt{positive}$ iff $p_{k,i}$ lies on the target described by $Q$; a point on a nearby distractor is therefore \texttt{negative} even when it lies on a foreground object. In implementation we batch all selected points from an image into a single grammar-guided generation call, reducing labeling to one VLM call per image regardless of $N$; for $K{>}1$ predicted bboxes we issue one prompt per bbox and dispatch them as a single vLLM batch, so per-image latency stays bounded by one VLM round-trip independent of $K$ (full prompt and EBNF grammar in Appendix~\ref{app:prompts_labeling}). Verification helps when a point falls on a boundary, background, or distractor; it can also misfire on target-interior points, and the reported confidence is a noisy self-report rather than a calibrated probability~\citep{OverconfidenceVLM2024NAACL, CalibratingLLMConfidence2024EMNLP}, so labeling complements rather than replaces the proactive boundary avoidance built into \methodshort{} (quantified in Section~\ref{sec:experiments:ablations}).

\subsection{Filtering and segmentation}
\label{sec:method:segmentation}

\paragraph{Point filtering.} Before prompting SAM we filter the labeled points using a confidence threshold $\tau$ and a label allowlist $\mathcal{A} \subseteq \{\texttt{positive}, \texttt{negative}\}$:
\begin{equation}
    \mathcal{P}_k^{\star}
    \;=\;
    \{\, p_{k,i} \in \mathcal{P}_k :
    c_{k,i} \geq \tau
    \;\wedge\;
    \ell_{k,i} \in \mathcal{A}
    \,\},
    \qquad
    \boldsymbol{\ell}_k^{\star} \;=\; (\ell_{k,i})_{p_{k,i} \in \mathcal{P}_k^{\star}}.
    \label{eq:filter}
\end{equation}
$\mathcal{A}=\{\texttt{positive}\}$ gives a positive-only prompt; $\mathcal{A}=\{\texttt{positive},\texttt{negative}\}$ passes both target-supporting and target-suppressing points; $\tau{=}0$ disables confidence filtering. Unless otherwise stated we use $\tau{=}0.7$ and $\mathcal{A}{=}\{\texttt{positive},\texttt{negative}\}$. 

\paragraph{SAM prompting and mask merging.} For each retained bbox we run SAM once,
\begin{equation}
    M_k
    \;=\;
    \operatorname{SAM}\!\bigl(
    I,\, b_k,\, \mathcal{P}_k^{\star},\, \boldsymbol{\ell}_k^{\star}\bigr),
    \qquad M_k \in \{0,1\}^{H \times W},
    \label{eq:sam_call}
\end{equation}
and combine per-bbox masks pixelwise:
\begin{equation}
    M \;=\; \textstyle\bigvee_{k=1}^{K} M_k.
    \label{eq:merge}
\end{equation}
When the VLM returns no bboxes ($K{=}0$), we set $M=\mathbf{0}_{H\times W}$, which contributes zero intersection and full ground-truth area to the cIoU denominator.


\section{Experiments}
\label{sec:experiments}

\subsection{Experimental setup}
\label{sec:experiments:setup}

\paragraph{Implementation details.}
Unless otherwise stated, experiments use Qwen2.5-VL-7B~\citep{Qwen25VL2025} (served via vLLM~\citep{vllm2023sosp}) as the frozen VLM and SAM\,2-L~\citep{SAM2_2024} as the frozen segmenter\footnote{We refer to SAM\,2-L simply as ``SAM'' in prose throughout for readability.}, and report both \methodshort{} fusion modes: \methodshort\textsubscript{mult} (multiplicative consensus, Eq.~\ref{eq:consensus_mult}) and \methodshort\textsubscript{msm} (max-saliency-smooth, Eq.~\ref{eq:consensus_msm}). Both VLM stages use grammar-guided structured decoding with $Temperature{=}0.1$ and a fixed seed, so \methodshort{}'s end-to-end output is stable in practice; the matched-budget naive baseline (Table~\ref{tab:main_tf}) involves stochastic sampling and is averaged over three seeds. All inference uses a single NVIDIA A6000 GPU at engine batch size $1$. Verbatim prompts and EBNF grammars for the localization and labeling stages are reproduced in Appendices~\ref{app:prompts_localization}--\ref{app:prompts_labeling}; full hyperparameters are in Appendix~\ref{app:reproducibility}.

\paragraph{Datasets and Evaluation metrics.}
We evaluate on both referring and reasoning segmentation benchmarks. For referring segmentation, we use three standard benchmarks: \textbf{RefCOCO}, \textbf{RefCOCO+}~\citep{refcoco2014emnlp, refcocoplus2017cvpr}, and \textbf{RefCOCOg}~\citep{refcocog2016cvpr}. RefCOCO/+/g are built from MSCOCO images with natural-language referring expressions; RefCOCO+ excludes spatial-relation descriptions, while RefCOCOg contains longer and more complex expressions. For reasoning segmentation, we evaluate on \textbf{ReasonSeg}~\citep{LISA_2024_CVPR}, where the model must segment the target specified by an implicit or reasoning-intensive instruction rather than an explicit referring expression.
Following common practice, we report cumulative Intersection-over-Union (cIoU) on all benchmarks, and additionally report the per-sample averaged IoU (gIoU) on ReasonSeg\footnote{We follow the LISA/ReasonSeg convention in calling the per-sample averaged IoU ``gIoU''; the same quantity is sometimes denoted mIoU elsewhere in the segmentation literature.}:
\begin{equation*}
    \mathrm{cIoU}
    =
    \frac{\sum_{i=1}^{N} |P_i \cap G_i|}
         {\sum_{i=1}^{N} |P_i \cup G_i|},
    \qquad
    \mathrm{gIoU}
    =
    \frac{1}{N}\sum_{i=1}^{N}
    \frac{|P_i \cap G_i|}{|P_i \cup G_i|}.
\end{equation*}

Here, $P_i$ is the predicted binary mask, $G_i$ is the ground-truth mask, and $N$ is the number of samples.

\subsection{Comparisons}
\label{sec:experiments:comparisons}

\begin{table}[t]
\centering
\small
\caption{\textbf{Informative points close the prompt-ambiguity gap.}
Training-free cIoU on RefCOCO/+/g (higher is better); \best{best} and \sbest{second-best} per column.
All non-Oracle rows use the same frozen VLM boxes, VLM point labels, point budget ($N{=}5$), and single SAM forward pass; only the point selector changes.
SAM4MLLM$^{\dagger}$ denotes our training-free reproduction of SAM4MLLM's PQPP setting, which uses uniformly sampled interior points.
The $\Delta$ rows therefore measure the effect of point selection alone, showing that replacing naive points with \method{} points yields consistent gains.
}
\label{tab:main_tf}
\setlength{\tabcolsep}{4pt}
\begin{tabular}{l ccc ccc cc}
\toprule
\multirow{2}{*}{\textbf{Method} (cIoU $\uparrow$)} & \multicolumn{3}{c}{\textbf{RefCOCO}} & \multicolumn{3}{c}{\textbf{RefCOCO+}} & \multicolumn{2}{c}{\textbf{RefCOCOg}} \\
\cmidrule(lr){2-4} \cmidrule(lr){5-7} \cmidrule(lr){8-9}
 & val & testA & testB & val & testA & testB & val(U) & test(U) \\
\midrule


Bounding box only                            & 69.2          & 72.0           & 63.9          & 60.7          & 68.7           & 56.9          & 65.1          & 67.4          \\

SAM4MLLM$^{\dagger}$                            & 61.9          & 65.5           & 57.6          & 53.2          & 59.6           & 46.9          & 55.9          & 57.1          \\
\midrule
\textsc{\methodshort}\textsubscript{mult} & \best{75.9}  & \sbest{81.7} & \sbest{71.1} & \best{71.4}  & \sbest{76.9} & \sbest{62.5} & \sbest{72.3} & \sbest{69.0} \\
\textsc{\methodshort}\textsubscript{msm}  & \sbest{75.4} & \best{81.9}  & \best{72.4}  & \sbest{71.3} & \best{77.1}  & \best{62.8}  & \best{73.6}  & \best{71.2}  \\
\cmidrule(lr){1-9}
$\Delta$\textsubscript{mult} \emph{vs.\ SAM4MLLM$^{\dagger}$} & \dpos{+14.0} & \dpos{+16.2} & \dpos{+13.5} & \dpos{+18.2} & \dpos{+17.3} & \dpos{+15.6} & \dpos{+16.4} & \dpos{+11.9} \\
$\Delta$\textsubscript{msm}  \emph{vs.\ SAM4MLLM$^{\dagger}$} & \dpos{+13.5} & \dpos{+16.4} & \dpos{+14.8} & \dpos{+18.1} & \dpos{+17.5} & \dpos{+15.9} & \dpos{+17.7} & \dpos{+14.1} \\

\bottomrule
\end{tabular}
\end{table}

\begin{table}[t]
\centering
\small
\caption{\textbf{Comparison against training-required methods.} cIoU on RefCOCO/+/g and gIoU/cIoU on ReasonSeg test, grouped by training regime. The RL block shares our Qwen2.5-VL-7B + SAM~2-L stack and is the closest direct comparison. \emph{Empirical Upper Bound}: cIoU with both VLM stages replaced by oracles. \best{Best} and \sbest{second-best} per column (excluding Oracle); baseline numbers are from the original publications, matched to our stack where reported or to the closest 7B-scale variant otherwise; absent cells (---) were not reported in the original paper.}
\label{tab:main}
\setlength{\tabcolsep}{3pt}
\begin{tabular}{l ccc ccc cc cc}
\toprule
\multirow{2}{*}{\textbf{Method}} & \multicolumn{3}{c}{\textbf{RefCOCO}} & \multicolumn{3}{c}{\textbf{RefCOCO+}} & \multicolumn{2}{c}{\textbf{RefCOCOg}} & \multicolumn{2}{c}{\textbf{ReasonSeg test}} \\
\cmidrule(lr){2-4} \cmidrule(lr){5-7} \cmidrule(lr){8-9} \cmidrule(lr){10-11}
 & val & testA & testB & val & testA & testB & val(U) & test(U) & gIoU & cIoU \\
\midrule

\multicolumn{11}{l}{\textit{Oracle reference}} \\
\emph{Empirical Upper Bound} & \emph{91.2} & \emph{91.2} & \emph{92.0} & \emph{91.2} & \emph{91.2} & \emph{92.1} & \emph{90.9} & \emph{90.1} & \emph{90.4} & \emph{90.6} \\

\midrule

\multicolumn{11}{l}{\textit{Supervised Fine-tuning}} \\
LISA-7B~\citep{LISA_2024_CVPR}                 & 74.9        & 79.1        & 72.3        & 65.1        & 70.8        & 58.1        & 67.9        & 70.6        & 48.7        & 48.8        \\
PixelLM-7B~\citep{PixelLM2024CVPR}             & 73.0        & 76.5        & 68.2        & 66.3        & 71.7        & 58.3        & 69.3        & 70.5        & ---         & ---         \\
VISA-7B~\citep{VISA2024ECCV}                   & 72.4        & 75.5        & 68.1        & 59.8        & 64.8        & 53.1        & 65.5        & 66.4        & ---         & ---         \\
PerceptionGPT-7B~\citep{PerceptionGPT2024CVPR} & ---         & 78.6        & ---         & ---         & 73.9        & ---         & ---         & 71.7        & ---         & ---         \\
SAM4MLLM~\citep{sam4mllm2024eccv}              & \best{77.1} & 80.9        & \best{72.5} & \best{71.5} & 76.8        & \best{64.7} & \best{74.5} & \best{75.2} & ---         & ---         \\

\midrule

\multicolumn{11}{l}{\textit{Reinforcement-Learning Tuning}} \\
Seg-Zero-7B~\citep{SegZero2025arxiv}            & ---         & 80.3         & ---         & ---         & 76.2        & ---         & ---         & 72.6         & 57.5         & 52.0         \\
SAM-R1-7B~\citep{SAMR1_2025}                    & ---         & 79.2         & ---         & ---         & 74.7        & ---         & ---         & 73.1         & 60.2         & 54.3         \\
SAM-Veteran-7B~\citep{SAMVeteran2025}           & ---         & 80.8         & ---         & ---         & 76.6        & ---         & ---         & \sbest{73.4} & \best{62.6}  & \best{56.1}  \\

\midrule

\multicolumn{11}{l}{\textit{Training-Free}} \\
\methodshort\textsubscript{mult} (ours) & \sbest{75.9} & \sbest{81.7} & 71.1         & \sbest{71.4} & \sbest{76.9} & 62.5         & 72.3         & 69.0 & \sbest{61.7} & \sbest{55.4} \\
\methodshort\textsubscript{msm} (ours)  & 75.4         & \best{81.9}  & \sbest{72.4} & 71.3         & \best{77.1}  & \sbest{62.8} & \sbest{73.6} & 71.2 & 60.9         & 54.9         \\

\bottomrule
\end{tabular}
\end{table}

We compare \methodshort{} with both training-free and training-required methods, with results summarized in Tables~\ref{tab:main_tf} and~\ref{tab:main}. Table~\ref{tab:main_tf} isolates the contribution of prompt quality against directly-related training-free baselines, and Table~\ref{tab:main} situates \methodshort{} within the broader literature dominated by training-required methods. Among training-free methods (Table~\ref{tab:main_tf}), both \methodshort{} variants substantially outperform naive point sampling at the same budget, with the dedicated $\Delta$ rows showing improvements of $+12$ to $+18$ cIoU across all eight RefCOCO/+/g splits. The only thing that differs between the SAM4MLLM$^{\dagger}$ row and the \methodshort{} rows is \emph{which} interior points are fed to SAM; the backbones, the bounding boxes, the budget, and the rest of the pipeline are identical, so the $\Delta$ rows cleanly isolate the contribution of prompt quality. Compared with training-required methods (Table~\ref{tab:main}), \methodshort{} is on par with fine-tuned and RL-tuned baselines without any task-specific training: it matches or exceeds fine-tuned baselines like LISA-7B and PixelLM-7B across most splits, and on RefCOCO \texttt{testA} reaches 81.9 cIoU, edging past all listed baselines including the SFT entry SAM4MLLM and the reinforcement-learning entries Seg-Zero and SAM-R1. The RL comparison is the cleanest one available, since Seg-Zero, SAM-R1, and SAM-Veteran use the same Qwen2.5-VL-7B + frozen SAM2 stack as \methodshort{} plus a few thousand GRPO training samples; \methodshort{} matches or beats Seg-Zero and SAM-R1 on both \texttt{testA} splits with no task-specific updates whatsoever, \emph{indicating that RL post-training to learn point selection is unnecessary on this stack.}
On the ReasonSeg test, \methodshort{} reaches $61.7/55.4$ (gIoU/cIoU), surpassing Seg-Zero ($57.5/52.0$) and SAM-R1 ($60.2/54.3$) and trailing the iterative agent SAM-Veteran ($62.6/56.1$) by under one cIoU point, again training-free.

\subsection{Ablation study}
\label{sec:experiments:ablations}

All ablations run on RefCOCO val with Qwen2.5-VL-7B + SAM\,2. The detailed end-to-end ablation table is reported in Appendix~\ref{app:ablations}; rows differ only in the point selector and the verifier configuration, with every other stage held fixed. Below, we summarize our key findings.

\paragraph{Hyperparameter sensitivity.} We sweep the three scalar hyperparameter classes whose effect on cIoU is non-trivial: the Gaussian-prior bandwidth $\sigma_G$, the edge/entropy penalty weights $(\lambda_E, \lambda_H)$ in $C_{\text{msm}}$, and the percentile-clip endpoints $(q_5, q_{95})$ (Appendix~\ref{app:reproducibility}). \methodshort{} is robust to moderate perturbations of any single knob; across every grid point, the cIoU on RefCOCO val and testB stays close to the default configuration, with $(\lambda_E, \lambda_H)$ the only axis where the choice meaningfully matters. The default $(\lambda_E, \lambda_H){=}(0.5, 1.5)$ achieves the highest mean cIoU across val and testB ($73.9$, vs. $73.2$ for the runner-up $(1.5, 0.5)$); it wins val by $1.2$ cIoU and trails the testB-best by only $0.5$, where the top three testB settings cluster within $0.5$ cIoU of each other. Holding $\lambda_E{=}0.5$ and increasing $\lambda_H$ from $0.5$ to $1.5$ improves val by $+2.4$ and testB by $+1.6$, suggesting $\lambda_H$ has a slightly stronger marginal effect than $\lambda_E$ inside $C_{\text{msm}}$, though the global landscape is non-monotonic and the two splits do not pick the same operating point.

\paragraph{Effect of filtering.}
Restricting $\mathcal{A}$ to a single label class is brittle: positive-only at $\tau{=}0.7$ gains $+0.4$ cIoU on val but loses $1.5$ on testB, and at $\tau{=}0.8$ the optimum flips again (positive-only wins on val, negative-only on testB), so the best label class is split-dependent. The threshold $\tau$ is more robust but bounded: verbalized confidences below ${\sim}0.7$ are systematically mis-labeled and benefit from removal, while above that cutoff confidence and correctness become uncorrelated, so pushing $\tau$ higher discards correct labels alongside wrong ones, and once the kept-prompt count drops the box-only fallback rate dominates ($-6.7$ cIoU; Table~\ref{tab:ablations}). At $\tau{=}0.9$ the both-labels cIoU sits $-0.8$ below default on both splits despite a $\sim\!45\%$ discard rate. We therefore keep $\tau{=}0.7,\,\mathcal{A}{=}\{\texttt{positive},\texttt{negative}\}$ as the most stable operating point; full grids in Appendix~\ref{app:ablations}.

\paragraph{Inference efficiency.}
\label{sec:experiments:efficiency}
Under our default configuration, a single query costs \emph{two VLM calls}, one for localization and one batched for point-labeling, \emph{independent of the point budget $N$, the number of proposed boxes $K$, and the choice of fusion mode}. This matches the inference-time call budget of RL-trained single-shot methods such as Seg-Zero~\citep{SegZero2025arxiv} and SAM-R1~\citep{SAMR1_2025}, and is strictly cheaper than decomposition pipelines~\citep{resanything2025neurips} or loop-based pipelines~\citep{SegLLM2024arxiv, CoT-RVS2025arxiv} whose call counts scale with expression length or refinement rounds. On RefCOCO val (single A6000) the two VLM calls together account for ${\sim}95\%$ of per-query wall-clock and \methodshort{}'s geometry-aware selector consumes only ${\sim}1\%$; the per-stage breakdown and a detailed call-count comparison are in Appendix~\ref{app:runtime}.

\begin{figure*}[t]
    \centering
    \subcaptionbox{%
        \textbf{Noisy or corrupted queries.}
        \label{fig:bench_quality_query_error}%
    }[0.485\textwidth]{%
        \includegraphics[width=0.485\textwidth]{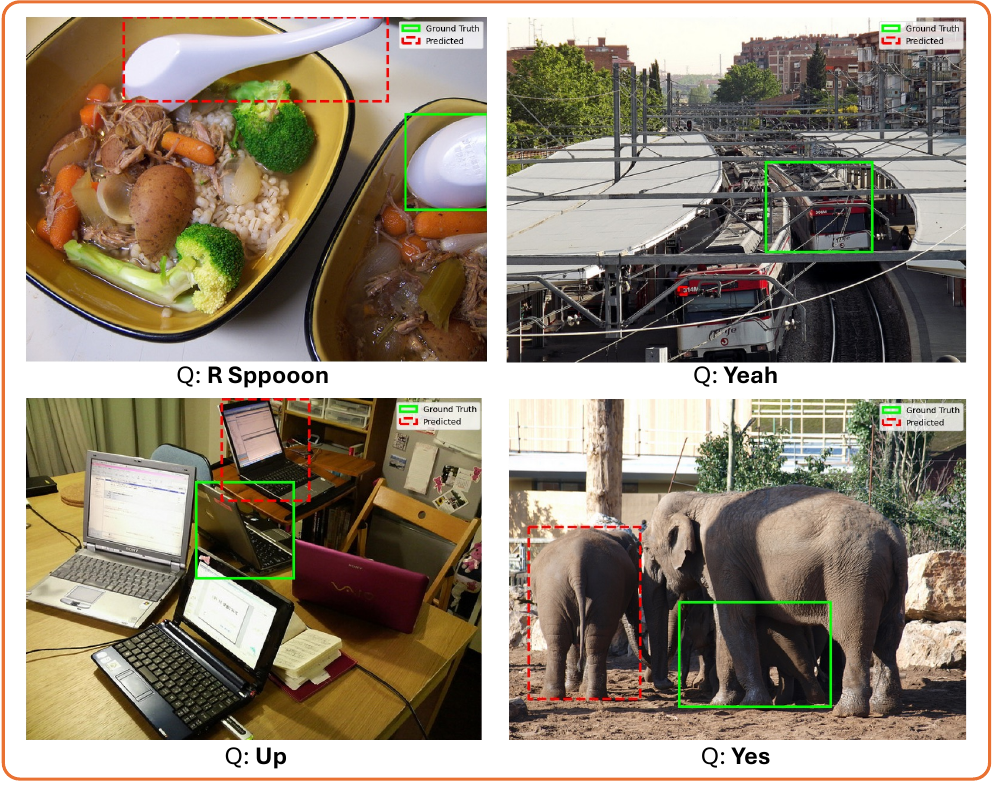}%
    }\hfill
    \subcaptionbox{%
        \textbf{Multiple valid targets.}
        \label{fig:bench_quality_ambiguous}%
    }[0.485\textwidth]{%
        \includegraphics[width=0.485\textwidth]{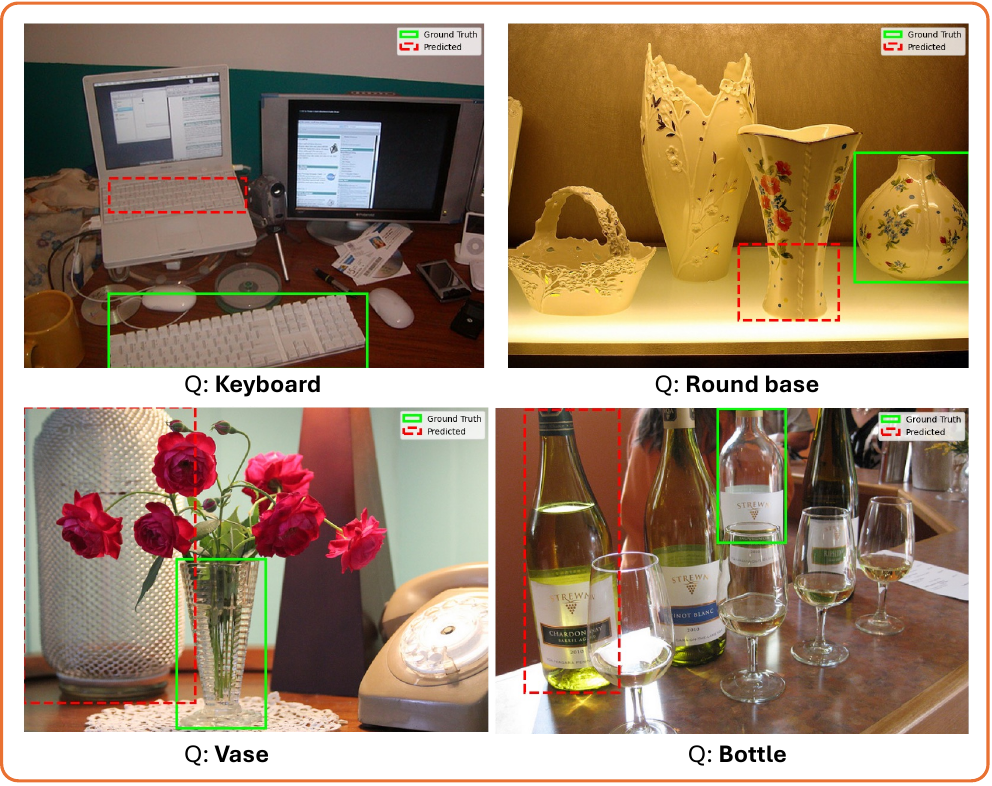}%
    }
    \caption{\textbf{Dataset noise and ambiguity limit the performance.} Even when a system selects a visually plausible region, evaluation can penalize it because the benchmark provides only one annotated target.  \textbf{(a)} Some queries are corrupted, unnatural, or under-specified, so the intended referent is not recoverable from language alone. \textbf{(b)} Other queries are valid but ambiguous: multiple objects satisfy the expression, while only one is labeled as ground truth.  Green boxes denote the benchmark target; red dashed boxes denote the predicted or plausible alternative target.}
    \label{fig:bench_quality_limits}
\end{figure*}

\paragraph{Benchmark noise and the oracle gap.}\label{sec:experiments:noise_and_oracle}
The empirical oracle reference in Table~\ref{tab:main} replaces VLM localization and point labeling with ground-truth-derived inputs while keeping \methodshort{} and SAM unchanged. The remaining gap to $100$\,cIoU therefore should not be interpreted as a VLM-grounding error; rather, it reflects the limits of SAM mask generation and the benchmark annotation protocol. \methodshort{}'s gap to this oracle reference primarily reflects the portion still attributable to imperfect localization and point verification. Figure~\ref{fig:bench_quality_limits} illustrates one residual source: some queries are corrupted, unnatural, or under-specified, making the intended referent ambiguous (Fig.~\ref{fig:bench_quality_query_error}), while others are valid but admit multiple plausible targets even though the benchmark provides only one (Fig.~\ref{fig:bench_quality_ambiguous}).

\section{Conclusion}
\label{sec:conclusion}
We argued that, in training-free VLM+SAM referring segmentation, the remaining gap to fine-tuned and RL-tuned specialists comes from \emph{prompt ambiguity} rather than from either backbone's capacity. Augmenting the VLM-predicted bounding box with semantically labeled points is a natural remedy. Still, the choice of points is decisive: poorly placed ones drift onto boundaries and distractors, and can degrade performance. The question that motivated this paper was whether selecting informative points requires learning at all.

Our answer is \methodshort{}: a deterministic method that fuses four classical cues into a consensus map, selects compact and spatially diverse candidates, and uses a frozen VLM as a per-point verifier. Because every component is frozen, \methodshort{} inherits their failure modes: the cue fusion biases against off-center, camouflaged, or heavily textured referents; and the VLM verifier remains systematically overconfident. Calibrating that confidence is a promising direction within the same pipeline.

Despite these limitations, \methodshort{} closes 12--18 cIoU over prior training-free methods at a matched budget on RefCOCO/+/g, and on the same stack as recent SFT and RL methods, it is competitive with them across RefCOCO/+/g and ReasonSeg at only two VLM calls per query. Learned point selection is therefore replaceable by image-side computation paired with semantic verification.

\begin{ack}
The authors would like to acknowledge the support from Prof. Sirisha Rambhatla’s Natural Sciences and Engineering Research Council of Canada (NSERC) Alliance Grant ALLRP 587270-2023.
\end{ack}

\bibliographystyle{plainnat}
\bibliography{references}

@InProceedings{LISA_2024_CVPR,
    author    = {Lai, Xin and Tian, Zhuotao and Chen, Yukang and Li, Yanwei and Yuan, Yuhui and Liu, Shu and Jia, Jiaya},
    title     = {{LISA}: Reasoning Segmentation via Large Language Model},
    booktitle = {Proceedings of the IEEE/CVF Conference on Computer Vision and Pattern Recognition (CVPR)},
    month     = {June},
    year      = {2024},
    pages     = {9579-9589}
}

@inproceedings{GRES2023CVPR, 
    title={{GRES}: Generalized Referring Expression Segmentation},
    url={http://dx.doi.org/10.1109/cvpr52729.2023.02259}, 
    DOI={10.1109/cvpr52729.2023.02259}, 
    booktitle={2023 IEEE/CVF Conference on Computer Vision and Pattern Recognition (CVPR)}, 
    publisher={IEEE}, 
    author={Liu, Chang and Ding, Henghui and Jiang, Xudong}, 
    year={2023}, 
    month=jun, 
    pages={23592–23601}
}

@misc{quantifyinglimitssegmentationfoundation2024arxiv,
      title={Quantifying the Limits of Segmentation Foundation Models: Modeling Challenges in Segmenting Tree-Like and Low-Contrast Objects}, 
      author={Yixin Zhang and Nicholas Konz and Kevin Kramer and Maciej A. Mazurowski},
      year={2025},
      eprint={2412.04243},
      archivePrefix={arXiv},
      primaryClass={cs.CV},
      url={https://arxiv.org/abs/2412.04243}, 
}

@inproceedings{sam4mllm2024eccv,
author = {Chen, Yi-Chia and Li, Wei-Hua and Sun, Cheng and Wang, Yu-Chiang Frank and Chen, Chu-Song},
title = {{SAM4MLLM}: Enhance Multi-Modal Large Language Model for Referring Expression Segmentation},
year = {2024},
isbn = {978-3-031-73003-0},
publisher = {Springer-Verlag},
address = {Berlin, Heidelberg},
url = {https://doi.org/10.1007/978-3-031-73004-7_19},
doi = {10.1007/978-3-031-73004-7_19},
booktitle = {Computer Vision – ECCV 2024: 18th European Conference, Milan, Italy, September 29–October 4, 2024, Proceedings, Part LXXXI},
pages = {323–340},
numpages = {18},
location = {Milan, Italy}
}

@article{PixelLM2024CVPR,
  title={{PixelLM}: Pixel Reasoning with Large Multimodal Model},
  author={Zhongwei Ren and Zhicheng Huang and Yunchao Wei and Yao Zhao and Dongmei Fu and Jiashi Feng and Xiaojie Jin},
  journal={2024 IEEE/CVF Conference on Computer Vision and Pattern Recognition (CVPR)},
  year={2024},
  pages={26364-26373},
  url={https://openaccess.thecvf.com/content/CVPR2024/html/Ren_PixelLM_Pixel_Reasoning_with_Large_Multimodal_Model_CVPR_2024_paper.html}
}

@article{SAMAug2024Arxiv,
  title={{SAMAug}: Point Prompt Augmentation for Segment Anything Model},
  author={Haixing Dai and Chong Ma and Zhiling Yan and Zhengliang Liu and Enze Shi and Yiwei Li and Peng Shu and Xiaozheng Wei and Lin Zhao and Zihao Wu and Fang Zeng and Dajiang Zhu and Wei Liu and Quanzheng Li and Lichao Sun and Shu Zhang and Tianming Liu and Xiang Li},
  journal={ArXiv},
  year={2023},
  volume={abs/2307.01187},
  url={https://arxiv.org/abs/2307.01187}
}

@InProceedings{GSVA2024CVPR,
    author    = {Xia, Zhuofan and Han, Dongchen and Han, Yizeng and Pan, Xuran and Song, Shiji and Huang, Gao},
    title     = {{GSVA}: Generalized Segmentation via Multimodal Large Language Models},
    booktitle = {Proceedings of the IEEE/CVF Conference on Computer Vision and Pattern Recognition (CVPR)},
    month     = {June},
    year      = {2024},
    pages     = {3858-3869}
}

@misc{thinkFirst2025Arxiv,
      title={Think Before You Segment: High-Quality Reasoning Segmentation with GPT Chain of Thoughts}, 
      author={Kao, Shiu-hong and Tai, Yu-Wing and Tang, Chi-Keung},
      year={2025},
      eprint={2503.07503},
      archivePrefix={arXiv},
      primaryClass={cs.CV},
      url={https://arxiv.org/abs/2503.07503}, 
}

@inproceedings{resanything2025neurips,
  title={{RESAnything}: Attribute Prompting for Arbitrary Referring Segmentation},
  author={Wang, Ruiqi and Zhang, Hao},
  booktitle={The Thirty-ninth Annual Conference on Neural Information Processing Systems (NeurIPS)},
  year={2025}
}

@misc{CoT-RVS2025arxiv,
      title={{CoT-RVS}: Zero-Shot Chain-of-Thought Reasoning Segmentation for Videos},
      author={Kao, Shiu-hong and Tai, Yu-Wing and Tang, Chi-Keung},
      year={2025},
      eprint={2505.18561},
      archivePrefix={arXiv},
      primaryClass={cs.CV},
      url={https://arxiv.org/abs/2505.18561}, 
}

@misc{SegLLM2024arxiv,
      title={{SegLLM}: Multi-round Reasoning Segmentation},
      author={XuDong Wang and Shaolun Zhang and Shufan Li and Konstantinos Kallidromitis and Kehan Li and Yusuke Kato and Kazuki Kozuka and Trevor Darrell},
      year={2024},
      eprint={2410.18923},
      archivePrefix={arXiv},
      primaryClass={cs.CV},
      url={https://arxiv.org/abs/2410.18923}, 
}

@inproceedings{SAM2023ICCV,
    author    = {Kirillov, Alexander and Mintun, Eric and Ravi, Nikhila and Mao, Hanzi and Rolland, Chloe and Gustafson, Laura and Xiao, Tete and Whitehead, Spencer and Berg, Alexander C. and Lo, Wan-Yen and Dollar, Piotr and Girshick, Ross},
    title     = {Segment Anything},
    booktitle = {Proceedings of the IEEE/CVF International Conference on Computer Vision (ICCV)},
    month     = {October},
    year      = {2023},
    pages     = {4015-4026}
}

@misc{SAM2_2024,
      title={{SAM 2}: Segment Anything in Images and Videos},
      author={Nikhila Ravi and Valentin Gabeur and Yuan-Ting Hu and Ronghang Hu and Chaitanya Ryali and Tengyu Ma and Haitham Khedr and Roman Radle and Chloe Rolland and Laura Gustafson and Eric Mintun and Junting Pan and Kalyan Vasudev Alwala and Nicolas Carion and Chao-Yuan Wu and Ross Girshick and Piotr Dollar and Christoph Feichtenhofer},
      year={2024},
      eprint={2408.00714},
      archivePrefix={arXiv},
      primaryClass={cs.CV},
      url={https://arxiv.org/abs/2408.00714},
}

@inproceedings{vllm2023sosp,
    title={Efficient Memory Management for Large Language Model Serving with {PagedAttention}},
    author={Woosuk Kwon and Zhuohan Li and Siyuan Zhuang and Ying Sheng and Lianmin Zheng and Cody Hao Yu and Joseph E. Gonzalez and Hao Zhang and Ion Stoica},
    booktitle={Proceedings of the ACM SIGOPS 29th Symposium on Operating Systems Principles},
    year={2023}
}

@inproceedings{refcoco2014emnlp,
    title={{ReferItGame}: Referring to Objects in Photographs of Natural Scenes},
    author={Kazemzadeh, Sahar and Ordonez, Vicente and Matten, Mark and Berg, Tamara},
    booktitle={Proceedings of the 2014 Conference on Empirical Methods in Natural Language Processing (EMNLP)},
    pages={787--798},
    year={2014}
}

@inproceedings{refcocog2016cvpr,
    title={Generation and Comprehension of Unambiguous Object Descriptions},
    author={Mao, Junhua and Huang, Jonathan and Toshev, Alexander and Camburu, Oana and Yuille, Alan L and Murphy, Kevin},
    booktitle={Proceedings of the IEEE Conference on Computer Vision and Pattern Recognition (CVPR)},
    pages={11--20},
    year={2016}
}

@inproceedings{refcocoplus2017cvpr,
    title={Modeling Context in Referring Expressions},
    author={Yu, Licheng and Poirson, Patrick and Yang, Shan and Berg, Alexander C. and Berg, Tamara L.},
    booktitle={European Conference on Computer Vision (ECCV)},
    pages={69--85},
    year={2016}
}

@inproceedings{soft-nms2017ICCV,
  title={{Soft-NMS}--improving object detection with one line of code},
  author={Bodla, Navaneeth and Singh, Bharat and Chellappa, Rama and Davis, Larry S},
  booktitle={Proceedings of the IEEE international conference on computer vision},
  pages={5561--5569},
  year={2017}
}

@inproceedings{StableSAM2025ICLR,
    title={Stable Segment Anything Model},
    author={Fan, Qi and Tao, Xin and Ke, Lei and Ye, Mingqiao and Zhang, Di and Wan, Pengfei and Tai, Yu-Wing and Tang, Chi-Keung},
    booktitle={International Conference on Learning Representations (ICLR)},
    year={2025},
    url={https://openreview.net/forum?id=ooxj2Audlq},
}

@inproceedings{OverconfidenceVLM2024NAACL,
    title={Overconfidence is Key: Verbalized Uncertainty Evaluation in Large Language and Vision-Language Models},
    author={Groot, Tobias and Valdenegro-Toro, Matias},
    booktitle={Proceedings of the 4th Workshop on Trustworthy Natural Language Processing (TrustNLP), NAACL},
    year={2024},
    url={https://arxiv.org/abs/2405.02917},
}

@inproceedings{CalibratingLLMConfidence2024EMNLP,
    title={Calibrating the Confidence of Large Language Models by Eliciting Fidelity},
    author={Zhang, Mozhi and Huang, Mianqiu and Shi, Rundong and Guo, Linsen and Peng, Chong and Yan, Peng and Zhou, Yaqian and Qiu, Xipeng},
    booktitle={Proceedings of the 2024 Conference on Empirical Methods in Natural Language Processing (EMNLP)},
    year={2024},
    url={https://aclanthology.org/2024.emnlp-main.173/},
}

@article{ittikoch1998,
    title={A model of saliency-based visual attention for rapid scene analysis},
    author={Itti, Laurent and Koch, Christof and Niebur, Ernst},
    journal={IEEE Transactions on Pattern Analysis and Machine Intelligence},
    volume={20},
    number={11},
    pages={1254--1259},
    year={1998},
    publisher={IEEE},
    doi={10.1109/34.730558}
}

@inproceedings{harelGBVS2006,
    title={Graph-based visual saliency},
    author={Harel, Jonathan and Koch, Christof and Perona, Pietro},
    booktitle={Advances in Neural Information Processing Systems (NeurIPS)},
    volume={19},
    pages={545--552},
    year={2006}
}

@inproceedings{hou2007spectralresidual,
    title={Saliency detection: A spectral residual approach},
    author={Hou, Xiaodi and Zhang, Liqing},
    booktitle={Proceedings of the IEEE Conference on Computer Vision and Pattern Recognition (CVPR)},
    pages={1--8},
    year={2007},
    doi={10.1109/CVPR.2007.383267}
}

@inproceedings{juddcenterbias2009,
    title={Learning to predict where humans look},
    author={Judd, Tilke and Ehinger, Krista and Durand, Fr{\'e}do and Torralba, Antonio},
    booktitle={Proceedings of the IEEE International Conference on Computer Vision (ICCV)},
    pages={2106--2113},
    year={2009},
    doi={10.1109/ICCV.2009.5459462}
}

@inproceedings{VISA2024ECCV,
  author    = {Cilin Yan and Haochen Wang and Shilin Yan and Xiaolong Jiang and
               Yao Hu and Guoliang Kang and Weidi Xie and Efstratios Gavves},
  editor    = {Ales Leonardis and Elisa Ricci and Stefan Roth and
               Olga Russakovsky and Torsten Sattler and G{\"u}l Varol},
  title     = {{VISA:} Reasoning Video Object Segmentation via Large Language Models},
  booktitle = {Computer Vision -- {ECCV} 2024 -- 18th European Conference, Milan, Italy,
               September 29--October 4, 2024, Proceedings, Part {XV}},
  series    = {Lecture Notes in Computer Science},
  pages     = {98--115},
  publisher = {Springer},
  year      = {2024},
  doi       = {10.1007/978-3-031-72633-0_6}
}

@inproceedings{PerceptionGPT2024CVPR,
  author    = {Renjie Pi and Lewei Yao and Jiahui Gao and Jipeng Zhang and Tong Zhang},
  title     = {{PerceptionGPT}: Effectively Fusing Visual Perception Into {LLM}},
  booktitle = {{IEEE/CVF} Conference on Computer Vision and Pattern Recognition,
               {CVPR} 2024, Seattle, WA, USA, June 16--22, 2024},
  pages     = {27114--27123},
  publisher = {{IEEE}},
  year      = {2024},
  doi       = {10.1109/CVPR52733.2024.02561}
}

@misc{SegZero2025arxiv,
  author        = {Yuqi Liu and Bohao Peng and Zhisheng Zhong and Zihao Yue and
                   Fanbin Lu and Bei Yu and Jiaya Jia},
  title         = {{Seg-Zero}: Reasoning-Chain Guided Segmentation via Cognitive Reinforcement},
  year          = {2025},
  eprint        = {2503.06520},
  archivePrefix = {arXiv},
  primaryClass  = {cs.CV},
  url           = {https://arxiv.org/abs/2503.06520}
}

@inproceedings{SAMR1_2025,
  title={{SAM}-R1: Leveraging {SAM} for Reward Feedback in Multimodal Segmentation via Reinforcement Learning},
  author={Jiaqi Huang and Zunnan Xu and Jun Zhou and Ting Liu and Yicheng Xiao and Mingwen Ou and Bowen Ji and Xiu Li and Kehong Yuan},
  booktitle={The Thirty-ninth Annual Conference on Neural Information Processing Systems},
  year={2026},
  url={https://openreview.net/forum?id=dHOSTp8MBl}
}

@inproceedings{SAMVeteran2025,
  author    = {Tianyuan Du and Haopeng Li and Zhen Fan and Jiarui Zhang and
               Panwang Pan and Yang Zhang},
  title     = {{SAM-Veteran}: An {MLLM}-Based Human-like {SAM} Agent for
               Reasoning Segmentation},
  booktitle = {The Fourteenth International Conference on Learning Representations (ICLR)},
  year      = {2026},
  url       = {https://openreview.net/forum?id=oN55r8iJJW}
}

@inproceedings{MaskRCNN2017ICCV,
  author    = {Kaiming He and Georgia Gkioxari and Piotr Doll\'{a}r and Ross Girshick},
  title     = {{Mask R-CNN}},
  booktitle = {Proceedings of the IEEE International Conference on Computer Vision (ICCV)},
  year      = {2017},
  url       = {https://arxiv.org/abs/1703.06870}
}

@inproceedings{Mask2Former2022CVPR,
  author    = {Bowen Cheng and Ishan Misra and Alexander G. Schwing and Alexander Kirillov
               and Rohit Girdhar},
  title     = {Masked-Attention Mask Transformer for Universal Image Segmentation},
  booktitle = {Proceedings of the IEEE/CVF Conference on Computer Vision and
               Pattern Recognition (CVPR)},
  year      = {2022},
  url       = {https://arxiv.org/abs/2112.01527}
}

@misc{MultimodalRISSurvey2025,
  author        = {Henghui Ding and Song Tang and Shuting He and Chang Liu and
                   Zuxuan Wu and Yu-Gang Jiang},
  title         = {Multimodal Referring Segmentation: A Survey},
  year          = {2025},
  eprint        = {2508.00265},
  archivePrefix = {arXiv},
  primaryClass  = {cs.CV},
  url           = {https://arxiv.org/abs/2508.00265}
}

@inproceedings{CogAgent2024CVPR,
  author    = {Wenyi Hong and Weihan Wang and Qingsong Lv and Jiazheng Xu and
               Wenmeng Yu and Junhui Ji and Yan Wang and Zihan Wang and
               Yuxuan Zhang and Juanzi Li and Bin Xu and Yuxiao Dong and
               Ming Ding and Jie Tang},
  title     = {{CogAgent}: A Visual Language Model for {GUI} Agents},
  booktitle = {Proceedings of the IEEE/CVF Conference on Computer Vision and
               Pattern Recognition (CVPR)},
  year      = {2024},
  url       = {https://arxiv.org/abs/2312.08914}
}

@inproceedings{InstructPix2Pix2023CVPR,
  author    = {Tim Brooks and Aleksander Holynski and Alexei A. Efros},
  title     = {{InstructPix2Pix}: Learning to Follow Image Editing Instructions},
  booktitle = {Proceedings of the IEEE/CVF Conference on Computer Vision and
               Pattern Recognition (CVPR)},
  year      = {2023},
  url       = {https://arxiv.org/abs/2211.09800}
}

@misc{InpaintAnything2023,
  author        = {Tao Yu and Runseng Feng and Ruoyu Feng and Jinming Liu and
                   Xin Jin and Wenjun Zeng and Zhibo Chen},
  title         = {Inpaint Anything: Segment Anything Meets Image Inpainting},
  year          = {2023},
  eprint        = {2304.06790},
  archivePrefix = {arXiv},
  primaryClass  = {cs.CV},
  url           = {https://arxiv.org/abs/2304.06790}
}

@inproceedings{LLaVA2023NeurIPS,
  author    = {Haotian Liu and Chunyuan Li and Qingyang Wu and Yong Jae Lee},
  title     = {Visual Instruction Tuning},
  booktitle = {Advances in Neural Information Processing Systems (NeurIPS)},
  year      = {2023},
  url       = {https://arxiv.org/abs/2304.08485}
}

@misc{Qwen25VL2025,
  author        = {Shuai Bai and Keqin Chen and Xuejing Liu and Jialin Wang and
                   Wenbin Ge and Sibo Song and Kai Dang and Peng Wang and Shijie Wang
                   and Jun Tang and Humen Zhong and Yuanzhi Zhu and Mingkun Yang and
                   Zhaohai Li and Jianqiang Wan and Pengfei Wang and Wei Ding and
                   Zheren Fu and Yiheng Xu and Jiabo Ye and Xi Zhang and Tianbao Xie
                   and Zesen Cheng and Hang Zhang and Zhibo Yang and Haiyang Xu and
                   Junyang Lin},
  title         = {{Qwen2.5-VL} Technical Report},
  year          = {2025},
  eprint        = {2502.13923},
  archivePrefix = {arXiv},
  primaryClass  = {cs.CV},
  url           = {https://arxiv.org/abs/2502.13923}
}

@misc{Qwen3VL2025,
  author        = {Shuai Bai and Yuxuan Cai and Keqin Chen and Junyang Lin and
                   Bowen Yu and Jingren Zhou and others},
  title         = {{Qwen3-VL} Technical Report},
  year          = {2025},
  eprint        = {2511.21631},
  archivePrefix = {arXiv},
  primaryClass  = {cs.CV},
  url           = {https://arxiv.org/abs/2511.21631}
}

@inproceedings{ProSAM2025ICCV,
  author    = {Wang, Xiaoqi and Sebastian, Clint and He, Wenbin and Ren, Liu},
  title     = {{ProSAM}: Enhancing the Robustness of {SAM}-based Visual Reference Segmentation with Probabilistic Prompts},
  booktitle = {Proceedings of the IEEE/CVF International Conference on Computer Vision (ICCV)},
  year      = {2025},
  url       = {https://arxiv.org/abs/2506.21835}
}

@inproceedings{LENS2026AAAI,
  author    = {Zhu, Lianghui and Ouyang, Bin and Zhang, Yuxuan and Cheng, Tianheng and Hu, Rui and Shen, Haocheng and Ran, Longjin and Chen, Xiaoxin and Yu, Li and Liu, Wenyu and Wang, Xinggang},
  title     = {{LENS}: Learning to Segment Anything with Unified Reinforced Reasoning},
  booktitle = {Proceedings of the AAAI Conference on Artificial Intelligence},
  year      = {2026},
  url       = {https://arxiv.org/abs/2508.14153}
}

@inproceedings{IteRPrimE2025AAAI,
  author    = {Wang, Yuji and Ni, Jingchen and Liu, Yong and Yuan, Chun and Tang, Yansong},
  title     = {{IteRPrimE}: Zero-shot Referring Image Segmentation with Iterative {Grad-CAM} Refinement and Primary Word Emphasis},
  booktitle = {Proceedings of the AAAI Conference on Artificial Intelligence},
  year      = {2025},
  url       = {https://ojs.aaai.org/index.php/AAAI/article/view/32880}
}

@misc{Sa2VA2025,
  author        = {Yuan, Haobo and Li, Xiangtai and Zhang, Tao and Sun, Yueyi and Huang, Zilong and Xu, Shilin and Ji, Shunping and Tong, Yunhai and Qi, Lu and Feng, Jiashi and Yang, Ming-Hsuan},
  title         = {{Sa2VA}: Marrying {SAM2} with {LLaVA} for Dense Grounded Understanding of Images and Videos},
  year          = {2025},
  eprint        = {2501.04001},
  archivePrefix = {arXiv},
  primaryClass  = {cs.CV},
  url           = {https://arxiv.org/abs/2501.04001}
}

@inproceedings{UniPixel2025NeurIPS,
  author    = {Liu, Ye and Ma, Zongyang and Pu, Junfu and Qi, Zhongang and Wu, Yang and Shan, Ying and Chen, Chang Wen},
  title     = {{UniPixel}: Unified Object Referring and Segmentation for Pixel-Level Visual Reasoning},
  booktitle = {Advances in Neural Information Processing Systems (NeurIPS)},
  year      = {2025},
  url       = {https://arxiv.org/abs/2509.18094}
}

\appendix


\section{\methodshort{} cue definitions, normalization, and Soft-NMS}
\label{app:cue_definitions}

This appendix gives the per-pixel formulas for the four cues used by \methodshort{}, the percentile-clipped normalization that puts them on a common scale, and the Soft-NMS recurrence used for diversity-aware point selection. All defaults are listed in Appendix~\ref{app:reproducibility}.

\paragraph{Color-contrast saliency.} $S$ rewards pixels whose local color distribution departs from their surroundings:
\begin{equation}
    S(p) \;=\; \lVert\, \mu^{\text{lab}}_{\text{in}}(p) \;-\; \mu^{\text{lab}}_{\text{ann}}(p) \,\rVert_2,
    \label{eq:cue_S}
\end{equation}
where $\mu^{\text{lab}}_{\text{in}}(p)$ and $\mu^{\text{lab}}_{\text{ann}}(p)$ are mean CIELAB colors in a local disk around $p$ and in the surrounding annulus, respectively. We work in CIELAB because it is approximately perceptually uniform: the $\ell_2$ distance between mean colors is only a meaningful contrast signal in a space where geometric distance tracks perceived color difference, which is not the case for RGB or HSV. CIELAB's perceptual uniformity is itself defined relative to a fixed reference illuminant (D65 in our implementation), and we do not perform white-balance or colour-constancy correction prior to computing $S$. Under strong shadows or specular highlights the $\ell_2$ contrast can therefore over- or under-estimate true perceptual difference, so we treat $S$ as a robust-but-approximate cue rather than a calibrated photometric quantity.

\paragraph{Edge density.} $E$ flags boundary-adjacent pixels, precisely the regions where SAM's decoder is most uncertain:
\begin{equation}
    E(p) \;=\; \bigl(K_{\text{box}} * \lVert \nabla I_{\text{gray}} \rVert \bigr)(p),
    \label{eq:cue_E}
\end{equation}
where ``$*$'' denotes 2D convolution and $K_{\text{box}}$ is a box-average kernel applied to the Sobel gradient magnitude $\lVert \nabla I_{\text{gray}} \rVert$. Prior to gradient computation, $I_{\text{gray}}$ is convolved with a Gaussian of $\sigma_{\text{pre}} = 1$ pixel to suppress high-frequency noise that would otherwise inflate $E$ (and $H$) on flat regions.

\paragraph{Local Shannon entropy.} $H$ measures textural clutter: a homogeneous neighborhood scores low, a busy one scores high:
\begin{equation}
    H(p) \;=\; -\!\sum_{v \in \mathcal{V}} \widehat{q}_v(p)\, \log \widehat{q}_v(p),
    \label{eq:cue_H}
\end{equation}
where $\widehat{q}_v(p)$ is the empirical frequency of quantized gradient-magnitude bin $v$ over a disk-shaped neighborhood of $p$, and $\mathcal{V}$ is a set of $|\mathcal{V}|\!=\!256$ equal-width bins. Concretely, we compute the per-pixel gradient magnitude $\lVert \nabla I_{\text{gray}} \rVert$, clip it to its $5$th and $95$th percentiles over the extended crop, rescale the clipped values linearly to the 8-bit integer range $[0, 255]$, and compute a $256$-level histogram over each disk-shaped neighborhood. The fixed $(5, 95)$ percentile clip used here is for entropy quantization and is independent of the percentile-clip endpoints $(q_5, q_{95})$ used in the cue normalizer (Eq.~\ref{eq:norm}); both happen to default to the same percentiles. Like $E$, $H$ identifies regions where both SAM's decoder and the VLM labeler are least reliable.

\paragraph{Gaussian spatial prior.} $G$ softly favors central positions inside the box:
\begin{equation}
    G(p) \;=\; \exp\!\Bigl(-\tfrac{\lVert p - c_b \rVert^2}{2\, \sigma_G^2}\Bigr),
    \qquad \sigma_G = 0.25\, d_{\text{diag}}(b),
    \label{eq:cue_G}
\end{equation}
where $c_b$ and $d_{\text{diag}}(b)$ are the center and the diagonal length of the bounding box $b$, and $\sigma_G$ is the Gaussian-prior bandwidth. $G$ acts as a weak geometric prior: it biases the consensus toward the interior without forbidding peripheral points.

\paragraph{Normalization.} The four raw cue maps have different scales, so we normalize each map to $[0,1]$ using percentile-clipped min--max scaling:
\begin{equation}
    \mathcal{N}(X)(p) \;=\; \operatorname{clip}\!\left(\frac{X(p) \;-\; q_{5}(X)}{q_{95}(X) \;-\; q_{5}(X) \;+\; \delta},\; 0,\; 1\right),
    \label{eq:norm}
\end{equation}
where $q_{\alpha}(X)$ is the $\alpha$-th percentile of $X$ over the extended bounding box and $\delta = 10^{-8}$ is a small numerical constant that prevents division by zero on degenerate (constant-valued) maps. We reuse $S,E,H,G$ to denote the normalized cues. Since high $E$ and high $H$ indicate unstable regions, the fusion rules in Section~\ref{sec:method:point_selection} use $(1-E)$ and $(1-H)$ as smoothness/homogeneity evidence to be rewarded.

\paragraph{Soft-NMS diversity selection.} Given a partially-selected set $\mathcal{P} \subseteq \widetilde{\mathcal{P}}$, each remaining candidate $p \in \widetilde{\mathcal{P}} \setminus \mathcal{P}$ receives the virtual score
\begin{equation}
    \widetilde{C}_{\mathcal{P}}(p)
    \;=\;
    C(p)\,
    \left(
        1 - \exp\!\left( -\frac{\min_{p' \in \mathcal{P}} \lVert p - p' \rVert^2}{2\sigma_{\text{nms}}^2} \right)
    \right),
    \label{eq:soft_nms}
\end{equation}
with the convention $\min_{p' \in \emptyset}\lVert p - p' \rVert^2 = +\infty$ so that $\widetilde{C}_{\emptyset}(p) = C(p)$ and the first selection reduces to $\arg\max_{p \in \widetilde{\mathcal{P}}} C(p)$. Subsequent points are chosen by $p^\star = \arg\max_{p} \widetilde{C}_{\mathcal{P}}(p)$. The selection terminates when $|\mathcal{P}| = N$, when $\widetilde{\mathcal{P}} \setminus \mathcal{P} = \emptyset$, or when every remaining virtual score falls below the floor $\varepsilon$ (early termination prevents \methodshort{} from padding the prompt with weak or redundant points). Algorithm~\ref{alg:phips} in the main text is the operational form of this recurrence.

\paragraph{Complexity.} Each cue map is computed over the extended crop, so cue extraction is $\mathcal{O}(|b_{\text{ext}}|)$ with small constants from box filtering, Sobel gradients, and local histograms. Local-maxima extraction is also $\mathcal{O}(|b_{\text{ext}}|)$, and Soft-NMS~\citep{soft-nms2017ICCV} selection over the candidate pool costs $\mathcal{O}(N \cdot |\widetilde{\mathcal{P}}|)$. \methodshort{} is deterministic image processing and adds no VLM calls beyond the localization and point-labeling stages; in practice, its cost is dominated by the surrounding VLM and SAM forward passes.

\section{Reproducibility details}
\label{app:reproducibility}

This appendix lists the default values for every symbol introduced in Section~\ref{sec:method}. All \methodshort{} radii are expressed relative to the diagonal length $d_{\text{diag}}(b)$ of the VLM-proposed bounding box so that the selector is scale-invariant across images and objects. Unless stated otherwise, these are the values used throughout every experiment in Section~\ref{sec:experiments}.

\begin{table}[h]
    \centering
    \small
    \setlength{\tabcolsep}{3pt}
    \caption{Default hyperparameters for \methodshort{}, point labeling, filtering, and SAM. Radii marked with $d_{\text{diag}}(b)$ scale with the per-box diagonal; fixed values are constants across all boxes and images.}
    \label{tab:reproducibility}
    \begin{tabular}{@{}l l l l@{}}
        \toprule
        \textbf{Symbol} & \textbf{Meaning} & \textbf{Default} & \textbf{Reference} \\
        \midrule
        $\alpha$                        & Box-extension factor for cue-map ROI & $0.20$                       & \S\ref{sec:method:point_selection} \\
        $r_{\text{in}}$                 & Saliency inner (local) window radius & $0.03 \cdot d_{\text{diag}}(b)$ & Eq.~\ref{eq:cue_S} \\
        $r_{\text{ann}}$                & Saliency annular window radius & $0.09 \cdot d_{\text{diag}}(b)$ & Eq.~\ref{eq:cue_S} \\
        $|K_{\text{box}}|$              & Edge-density box-average kernel side & $0.05 \cdot d_{\text{diag}}(b)$ & Eq.~\ref{eq:cue_E} \\
        $\sigma_{\text{pre}}$           & Pre-Sobel Gaussian smoothing & $1.0$ pixel                  & Eq.~\ref{eq:cue_E} \\
        $r_{H}$                         & Entropy disk-neighborhood radius & $0.03 \cdot d_{\text{diag}}(b)$ & Eq.~\ref{eq:cue_H} \\
        $|\mathcal{V}|$                 & Gradient-magnitude quantization bins & $256$ (8-bit)               & Eq.~\ref{eq:cue_H} \\
        $\sigma_G$                      & Gaussian-prior bandwidth & $0.25 \cdot d_{\text{diag}}(b)$ & Eq.~\ref{eq:cue_G} \\
        $(q_5, q_{95})$                 & Cue-normalizer percentile-clip endpoints & $(5\text{th},\, 95\text{th})$ & Eq.~\ref{eq:norm} \\
        $\delta$                        & Numerical-stability constant in $\mathcal{N}(\cdot)$ & $10^{-8}$ & Eq.~\ref{eq:norm} \\
        $(p^{H}_{\text{lo}}, p^{H}_{\text{hi}})$ & Gradient-mag.\ clip for $H$ quantization & $(5\text{th},\, 95\text{th})$ & Eq.~\ref{eq:cue_H} \\
        $\lambda_E$                     & Edge penalty in $C_{\text{msm}}$ & $0.5$                        & Eq.~\ref{eq:consensus_msm} \\
        $\lambda_H$                     & Entropy penalty in $C_{\text{msm}}$ & $1.5$                        & Eq.~\ref{eq:consensus_msm} \\
        $W_{\text{peak}}$               & Local-maxima peak-finding window & $3 \times 3$, 8-connected     & \S\ref{sec:method:point_selection} \\
        $\sigma_{\text{nms}}$           & Soft-NMS Gaussian suppression radius & $0.10 \cdot d_{\text{diag}}(b)$ & Eq.~\ref{eq:soft_nms} \\
        $\varepsilon$                   & Soft-NMS termination floor & $10^{-3}$                    & \S\ref{sec:method:point_selection} \\
        $N$                             & Requested point budget per box & $5$                          & \S\ref{sec:method:point_selection} \\
        $\tau$                          & Confidence threshold for filtering & $0.7$                        & Eq.~\ref{eq:filter} \\
        $\mathcal{A}$                   & Label allowlist & $\{\texttt{positive},\, \texttt{negative}\}$ & Eq.~\ref{eq:filter} \\
        \midrule
        \multicolumn{2}{@{}l}{\textbf{VLM decoding}} & & \\
        Temperature                     & Sampling temperature & $0.1$                        & \S\ref{sec:method:point_labeling} \\
        top-$p$                         & Nucleus sampling & $0.9$                        & \S\ref{sec:method:point_labeling} \\
        max-new-tokens                  & Generation budget per call & $512$                        & \S\ref{sec:method:point_labeling} \\
        Labeling batching               & All $\mathcal{P}_k$ batched; $\sim$2 calls/image & on & \S\ref{sec:method:point_labeling} \\
        \midrule
        \multicolumn{2}{@{}l}{\textbf{SAM}} & & \\
        Checkpoint                      & SAM\,2 model identifier (HF Hub) & \texttt{facebook/sam2-hiera-large} & \S\ref{sec:method:segmentation} \\
        \texttt{multimask\_output}      & Mask proposals per prompt & \texttt{False} (single-mask) & Eq.~\ref{eq:sam_call} \\
        \bottomrule
    \end{tabular}
\end{table}

\paragraph{Deterministic behavior.}
Every stage of \methodshort{} is deterministic for a fixed image, box, and budget. The only stochastic component is VLM sampling in the localization and point-labeling stages; we use a low temperature ($0.1$) and a fixed nucleus ($0.9$) so that the structured JSON outputs are stable in practice, and every benchmark number reported in Section~\ref{sec:experiments} is a single pass with a fixed seed.

\paragraph{Hardware.}
All inference runs use a single NVIDIA A6000 GPU with batch size $1$ at the engine level; vLLM internally batches the per-image point-labeling calls. The two-VLM-call cost structure is hardware-independent; see Section~\ref{sec:experiments:efficiency} for the call-count analysis.

\paragraph{Hyperparameter sensitivity.}
Table~\ref{tab:phips_sensitivity} sweeps the three scalar hyperparameter classes whose effect on cIoU is non-trivial: the Gaussian-prior bandwidth $\sigma_G$ (Eq.~\ref{eq:cue_G}), the edge / entropy penalty weights $(\lambda_E, \lambda_H)$ in $C_{\text{msm}}$ (Eq.~\ref{eq:consensus_msm}), and the percentile-clip endpoints $(q_5, q_{95})$ in the cue normalizer (Eq.~\ref{eq:norm}). Each block sweeps one parameter while holding the remaining defaults from Table~\ref{tab:reproducibility} fixed; the reported cIoU values are based on runs on RefCOCO val and the more spatially-challenging testB split. Three observations follow. First, $\sigma_G$ is essentially flat across the reasonable range $[0.10, 0.33]$ (val cIoU spans only $1.1$ points), with a meaningful drop only at the over-smoothed endpoint $\sigma_G{=}0.45$; the default $0.25$ is robust. Second, the $(\lambda_E, \lambda_H)$ grid is the most informative axis but the two splits do not agree on a single winner: $(0.5, 1.5)$ wins val with $75.4$ cIoU and reaches $72.4$ on testB, while $(1.5, 0.5)$ tops testB with $72.9$ but loses $1.9$ cIoU on val. We adopt $(0.5, 1.5)$ as the default reflected in Table~\ref{tab:reproducibility} because (i) it wins val by a clear margin and is within $0.5$ cIoU of the best on testB (where the top settings cluster within a narrow band of $0.5$ cIoU), and (ii) holding $\lambda_E{=}0.5$ fixed and sweeping $\lambda_H \in \{0.5, 1.0, 1.5\}$ yields a monotonic gain on both splits ($+2.4$ val, $+1.6$ testB), indicating that within the smallest-edge-penalty slice the entropy penalty does have a slightly stronger marginal effect; the global $3{\times}3$ landscape is, however, non-monotonic, so this should be read as a local rather than universal trend. Third, the percentile-clip endpoints $(q_5, q_{95})$ are stable: the wider $(1, 99)$ clip edges $(5, 95)$ on val by $0.4$ cIoU but loses $1.6$ cIoU on testB, so we keep the more stable $(5, 95)$ default. Across every grid point the cIoU stays close to the default configuration (highlighted in bold), confirming that \methodshort{} does not depend on a brittle hyperparameter setting.

\begin{table}[h]
    \centering
    \small
    \setlength{\tabcolsep}{4pt}
    \caption{\textbf{Hyperparameter sensitivity.} cIoU on RefCOCO val and testB while sweeping $\sigma_G$ (as a fraction of $d_{\text{diag}}(b)$), $(\lambda_E, \lambda_H)$, and $(q_5, q_{95})$; $\Delta$ columns report the gap to the block's default row. \best{Best} and \sbest{second-best} per metric per block; the default row labels are bolded. The $(\lambda_E, \lambda_H)$ block is reported under $C_{\text{msm}}$ since these weights only appear in Eq.~\ref{eq:consensus_msm}; the $\sigma_G$ and $(q_5, q_{95})$ blocks are reported under the paper-default $C_{\text{mult}}$, which is why the ``default'' rows do not all share the same cIoU value.}
    \label{tab:phips_sensitivity}
    \begin{tabular}{@{}l l c c c c@{}}
        \toprule
        \textbf{Parameter} & \textbf{Setting} & \textbf{val cIoU} $\uparrow$ & \textbf{$\Delta$ val} & \textbf{testB cIoU} $\uparrow$ & \textbf{$\Delta$ testB} \\
        \midrule
        \multicolumn{6}{@{}l}{\emph{Gaussian-prior bandwidth $\sigma_G$ (frac.\ of $d_{\text{diag}}(b)$)}} \\
        $\sigma_G$ & $0.10$ & \sbest{$75.9$} & $0.0$         & $69.9$         & \dneg{$-1.2$} \\
        $\sigma_G$ & $0.18$ & \best{$76.6$}  & \dpos{$+0.7$} & \sbest{$70.8$} & \dneg{$-0.3$} \\
        $\boldsymbol{\sigma_G}$ & $\boldsymbol{0.25}$ \emph{(default)} & \sbest{$\boldsymbol{75.9}$} & $\boldsymbol{0.0}$ & \best{$\boldsymbol{71.1}$}  & $\boldsymbol{0.0}$ \\
        $\sigma_G$ & $0.33$ & $75.5$ & \dneg{$-0.4$} & $70.7$ & \dneg{$-0.4$} \\
        $\sigma_G$ & $0.45$ & $73.6$ & \dneg{$-2.3$} & $70.5$ & \dneg{$-0.6$} \\
        \midrule
        \multicolumn{6}{@{}l}{\emph{Edge / entropy penalty weights $(\lambda_E, \lambda_H)$ in $C_{\text{msm}}$}} \\
        $(\lambda_E, \lambda_H)$ & $(0.5, 0.5)$ & $73.0$         & \dneg{$-2.4$} & $70.8$         & \dneg{$-1.6$} \\
        $(\lambda_E, \lambda_H)$ & $(0.5, 1.0)$ & $72.6$         & \dneg{$-2.8$} & $71.0$         & \dneg{$-1.4$} \\
        $\boldsymbol{(\lambda_E, \lambda_H)}$ & $\boldsymbol{(0.5, 1.5)}$ \emph{(default)} & \best{$\boldsymbol{75.4}$} & $\boldsymbol{0.0}$ & $\boldsymbol{72.4}$ & $\boldsymbol{0.0}$ \\
        $(\lambda_E, \lambda_H)$ & $(1.0, 0.5)$ & \sbest{$74.2$} & \dneg{$-1.2$} & $70.0$         & \dneg{$-2.4$} \\
        $(\lambda_E, \lambda_H)$ & $(1.0, 1.0)$ & $73.0$         & \dneg{$-2.4$} & $70.5$         & \dneg{$-1.9$} \\
        $(\lambda_E, \lambda_H)$ & $(1.0, 1.5)$ & $73.0$         & \dneg{$-2.4$} & \sbest{$72.7$} & \dpos{$+0.3$} \\
        $(\lambda_E, \lambda_H)$ & $(1.5, 0.5)$ & $73.5$         & \dneg{$-1.9$} & \best{$72.9$}  & \dpos{$+0.5$} \\
        $(\lambda_E, \lambda_H)$ & $(1.5, 1.0)$ & $71.9$         & \dneg{$-3.5$} & $71.4$         & \dneg{$-1.0$} \\
        $(\lambda_E, \lambda_H)$ & $(1.5, 1.5)$ & $73.1$         & \dneg{$-2.3$} & $69.6$         & \dneg{$-2.8$} \\
        \midrule
        \multicolumn{6}{@{}l}{\emph{Percentile-clip endpoints $(q_5, q_{95})$ in the cue normalizer}} \\
        $(q_5, q_{95})$ & $(1, 99)$ & \best{$76.3$}  & \dpos{$+0.4$} & $69.5$         & \dneg{$-1.6$} \\
        $\boldsymbol{(q_5, q_{95})}$ & $\boldsymbol{(5, 95)}$ \emph{(default)} & \sbest{$\boldsymbol{75.9}$} & $\boldsymbol{0.0}$ & \best{$\boldsymbol{71.1}$} & $\boldsymbol{0.0}$ \\
        $(q_5, q_{95})$ & $(10, 90)$ & $75.6$ & \dneg{$-0.3$} & \sbest{$69.6$} & \dneg{$-1.5$} \\
        \bottomrule
    \end{tabular}
\end{table}

\section{Localization prompt and grammar}
\label{app:prompts_localization}

The first stage of the pipeline uses the VLM as a grounded proposer: given an image $I$ and a natural-language query $Q$, it returns a (possibly empty) set of candidate bounding boxes. To make the output consumable by downstream stages we constrain the generation with a formal EBNF grammar via vLLM's structured outputs. The prompt template below is what the VLM sees; the grammar enforces that its output is a single valid JSON object with chain-of-thought and box coordinates.

\begin{tcolorbox}[
    colback=gray!5,
    colframe=black!55,
    title=\textbf{Localization Prompt},
    fonttitle=\bfseries,
    boxrule=0.5pt,
    arc=2pt,
    breakable
]
\small\ttfamily
You are a visual grounding assistant.\\[2pt]
Given an image and a natural-language query, return every object in the image\\
that the query refers to, as a bounding box in pixel coordinates.\\[2pt]
Image size: \{H\} $\times$ \{W\} (height $\times$ width).\\
Query: ``\{Q\}''\\[2pt]
Respond with a single JSON object that conforms to the grammar below.\\
Include a short \texttt{reasoning} string explaining your choice, followed by a\\
\texttt{boxes} list in which each element has integer fields\\
\texttt{x\_min, y\_min, x\_max, y\_max}, all in pixel coordinates of the image.\\
If the query does not refer to any visible object, return an empty list for\\
\texttt{boxes} and briefly explain why in \texttt{reasoning}.
\end{tcolorbox}

\begin{tcolorbox}[
    colback=gray!5,
    colframe=black!55,
    title=\textbf{EBNF Grammar (Localization)},
    fonttitle=\bfseries,
    boxrule=0.5pt,
    arc=2pt,
    breakable
]
\small
\begin{verbatim}
root        ::= "{" ws "\"reasoning\"" ws ":" ws string
                    ws "," ws "\"boxes\"" ws ":" ws box_list
                    ws "}"

box_list    ::= "[" ws "]" | "[" ws box (ws "," ws box)* ws "]"

box         ::= "{" ws "\"x_min\"" ws ":" ws integer
                    ws "," ws "\"y_min\"" ws ":" ws integer
                    ws "," ws "\"x_max\"" ws ":" ws integer
                    ws "," ws "\"y_max\"" ws ":" ws integer
                    ws "}"

integer     ::= "0" | [1-9] [0-9]*
string      ::= "\"" ( [^"\\] | "\\" . )* "\""
ws          ::= ( " " | "\t" | "\n" | "\r" )*
\end{verbatim}
\end{tcolorbox}

\paragraph{Implementation note.}
The grammar is enforced at decoding time via vLLM's \texttt{StructuredOutputsParams}. If the grammar-constrained decoder truncates early (e.g., the generation budget is exhausted mid-JSON), we fall back to a single unconstrained retry and discard the sample only if the retry also fails to produce a parseable object; empirically this happens on fewer than $0.1\%$ of the RefCOCO/+/g validation samples with the default \texttt{max-new-tokens} budget.

\section{Point-labeling prompt and grammar}
\label{app:prompts_labeling}

The third stage of the pipeline uses the VLM as a semantic verifier: for each \methodshort{}-selected point $p_{k,i} \in \mathcal{P}_k$ inside the proposed box $b_k$, the VLM returns a label $\ell_{k,i} \in \{\texttt{positive}, \texttt{negative}\}$ and a verbalized confidence $c_{k,i} \in [0,1]$. As discussed in Section~\ref{sec:method:point_labeling}, the VLM receives the image, the query, the box, and the point coordinates via text; no pixel-level annotations (dots, crosses, rectangles) are rendered onto the image before it is shown to the VLM. All points inside an image are labeled in a single batched generation call.

\begin{tcolorbox}[
    colback=gray!5,
    colframe=black!55,
    title=\textbf{Point-Labeling Prompt},
    fonttitle=\bfseries,
    boxrule=0.5pt,
    arc=2pt,
    breakable
]
\small\ttfamily
You are a visual grounding assistant.\\[2pt]
Given an image, a natural-language query, a bounding box, and a list of points\\
inside that bounding box, decide for each point whether it lies on the object\\
described by the query (\texttt{positive}) or on the background (\texttt{negative}),\\
and report your confidence.\\[2pt]
Image size: \{H\} $\times$ \{W\} (height $\times$ width).\\
Query: ``\{Q\}''\\
Bounding box (pixel coordinates): \{x\_min, y\_min, x\_max, y\_max\}.\\
Points: [\{x\_1, y\_1\}, \{x\_2, y\_2\}, \ldots, \{x\_n, y\_n\}].\\[2pt]
Respond with a single JSON object that conforms to the grammar below.\\
The \texttt{labels} field must contain exactly one entry per input point, in the\\
same order. Each entry has a \texttt{label} string (\texttt{positive} or\\
\texttt{negative}) and a \texttt{confidence} number in $[0, 1]$.\\
Do not hedge: pick the label that better describes the point, and report your\\
own honest confidence.
\end{tcolorbox}

\begin{tcolorbox}[
    colback=gray!5,
    colframe=black!55,
    title=\textbf{EBNF Grammar (Point Labeling)},
    fonttitle=\bfseries,
    boxrule=0.5pt,
    arc=2pt,
    breakable
]
\small
\begin{verbatim}
root        ::= "{" ws "\"labels\"" ws ":" ws label_list ws "}"

label_list  ::= "[" ws label (ws "," ws label)* ws "]"

label       ::= "{" ws "\"label\"" ws ":" ws label_enum
                    ws "," ws "\"confidence\"" ws ":" ws confidence
                    ws "}"

label_enum  ::= "\"positive\"" | "\"negative\""

confidence  ::= "0" | "1" | "0." [0-9]+ | "1.0" | "1.00"

ws          ::= ( " " | "\t" | "\n" | "\r" )*
\end{verbatim}
\end{tcolorbox}

\paragraph{Batching.}
In practice, all points for a single image (and, when the benchmark is run in batched mode, all points across a minibatch of images) are passed to the VLM in one structured-output call. This collapses point labeling from $|\mathcal{P}_k|$ sequential forward passes to a single forward pass per image, and the whole pipeline consumes approximately two VLM calls per image in total: one for localization and one for the entire labeling batch. The template above handles one box per call; multi-box images dispatch $K$ such prompts concurrently as a single vLLM batch, so per-image latency stays bounded by one VLM round-trip independent of $K$.

\paragraph{Length-mismatch fallback.}
The EBNF grammar enforces the JSON structure of the response but does not constrain the \texttt{labels} array length to match the input point count. We handle the two failure modes deterministically: if the VLM returns a \texttt{labels} array \emph{longer} than the input point list, we truncate to the input length; if it returns a \emph{shorter} array, the missing trailing entries are treated as low-confidence labels and are dropped by the confidence filter (Eq.~\ref{eq:filter}). No retry is issued, and length mismatches do not propagate to the SAM call.

\section{End-to-end ablation}
\label{app:ablations}

This appendix reports the detailed end-to-end ablation summarized in Section~\ref{sec:experiments:ablations}. Each row holds every stage of the \methodshort{} pipeline fixed except the named factor; $N{=}5$ throughout.

\begin{table}[h]
\centering
\small
\setlength{\tabcolsep}{6pt}
\caption{\textbf{End-to-end ablation on RefCOCO val.} Each row holds every stage of the pipeline fixed except the named factor; $N{=}5$ throughout. ``Random'' samples points uniformly inside the VLM box (averaged over three seeds); ``All-positive'' skips the VLM verifier and labels every \methodshort{} point as \texttt{positive}; ``$\tau{=}0$'' disables confidence filtering. Bold marks the default \methodshort{} configuration (also the best result) used throughout the rest of the paper; \sbest{underline} marks the second-best.}
\label{tab:ablations}
\begin{tabular}{@{}l l l c c@{}}
\toprule
\textbf{Selector} & \textbf{Verifier} & \textbf{Filter} & \textbf{cIoU} $\uparrow$ & $\Delta$ \emph{vs.\ default} \\
\midrule
Bounding box only                & ---                  & ---           & $69.2$         & \dneg{$-6.7$}  \\
Naive sampling                   & VLM labels           & $\tau{=}0.7$  & $61.9$         & \dneg{$-14.0$} \\
\methodshort\textsubscript{mult} & all-positive         & ---           & $61.1$         & \dneg{$-14.8$} \\
\methodshort\textsubscript{mult} & VLM labels           & $\tau{=}0$    & $73.6$         & \dneg{$-2.3$}  \\
\textbf{\methodshort\textsubscript{mult}} & \textbf{VLM labels} & $\boldsymbol{\tau{=}0.7}$ \emph{(default)} & $\boldsymbol{75.9}$ & $\boldsymbol{0.0}$ \\
\methodshort\textsubscript{msm}  & VLM labels           & $\tau{=}0.7$  & \sbest{$75.4$} & \dneg{$-0.5$} \\
\bottomrule
\end{tabular}
\end{table}

\paragraph{Roles of the verifier and the filter.}
The verifier and the confidence filter contribute distinct, complementary roles even though they both operate on the labeled point set. \methodshort{}'s point selection is semantically blind: the four cues observe the image and the bounding box, but never the referring expression. Treating every selected point as positive (the all-positive row) therefore makes the prompt more ambiguous, since any candidate that lands on co-occurring background or a distractor object pollutes the support; the VLM verifier converts these geometrically-chosen candidates into positive/negative SAM prompts that align with the referring expression. Numerically, dropping the verifier costs $14.8$ cIoU while disabling the threshold ($\tau{=}0$) costs only $2.3$ cIoU, so the verifier carries the bulk of the work and the filter acts as a stabilizer that discards low-confidence labels so occasional VLM misjudgments do not propagate into the SAM mask. The two fusion modes (last two rows) sit within $0.5$ cIoU of each other on this split, supporting our framing of $C_{\text{msm}}$ as a softer relaxation of $C_{\text{mult}}$ rather than a separate method.

\paragraph{Threshold sensitivity and VLM portability.}
Two questions follow from the $-2.3$ cIoU gap between $\tau{=}0$ and $\tau{=}0.7$ in the ablation above. First, is $\tau{=}0.7$ brittle on the VLM we use, or is the entire $[0, 0.7]$ range close to the default? The two endpoints give a $2.3$ cIoU envelope, and verbalized confidences from grammar-guided Qwen2.5-VL are concentrated above $0.5$ in our runs, so most of that envelope is contributed by a small slice of low-confidence labels rather than by a sharp peak around $0.7$; \methodshort{} is therefore not relying on a finely tuned cutoff for this VLM. Second, does the same value of $\tau$ port to a different VLM? Verbalized confidences are systematically overconfident and miscalibrated, with the degree of miscalibration differing meaningfully across language and vision--language models tested side by side~\citep{OverconfidenceVLM2024NAACL}, so the absolute scale of $c_{k,i}$ is VLM-specific and the optimal $\tau$ is unlikely to transfer numerically. The role of the threshold (suppress the lowest-confidence labels), however, is a structural property of the filter and not of the value itself: any VLM that reports a usable confidence score will benefit from removing its bottom slice, even if the cut-off needs to be re-located. We therefore recommend a coarse one-pass val sweep over $\tau \in \{0.0, 0.3, 0.5, 0.7, 0.9\}$ when porting \methodshort{} to a new VLM, rather than treating $0.7$ as a universal default; the verifier itself, the only stage with a large effect, is unchanged across this sweep.

\paragraph{Stage decomposition and per-cue robustness.}
Table~\ref{tab:ablations} holds the pipeline fixed at the default configuration and ablates one factor at a time; we complement it here with two analyses that probe different axes. Table~\ref{tab:stage_ablation} walks a candidate prompt set through five progressive pipeline configurations, from the box-only floor to the full \methodshort{} pipeline, isolating what each stage contributes on top of the previous one. Table~\ref{tab:cue_ablation} zeros each of the four \methodshort{} cues---colour-contrast saliency $S$, edge density $E$, local entropy $H$, and the Gaussian spatial prior $G$---one at a time, asking whether the gain is robust to dropping any single cue. Together the two answer two questions Table~\ref{tab:ablations} does not address by construction: \emph{where} along the pipeline the gain enters, and \emph{whether} the four-cue selector relies on any single cue. The cue ablation runs under a diagnostic linear-penalty fusion mode, $C(p) = S(p)\cdot\max(1-\lambda_E E(p) - \lambda_H H(p),\,0)\cdot G(p)$, which is the only fusion that admits a clean off-switch for $E$ and $H$ via the existing $\lambda_E$ and $\lambda_H$ knobs (the headline $C_{\text{mult}}$ multiplies $S$, $(1-E)$, $(1-H)$, $G$ with no scalar weights, so an off-switch for any single cue would change the formula rather than zero a parameter); we therefore report $\Delta$cIoU vs.\ the all-cues-on baseline rather than absolute cIoU. $S$ enters every fusion as a multiplicative factor and admits no zero-weight off-switch in any mode, so a $\text{no}\_S$ row is intentionally omitted.

\begin{table}[h]
\centering
\small
\setlength{\tabcolsep}{6pt}
\caption{\textbf{Per-stage contribution on RefCOCO val.} Five progressive configurations from the box-only floor to the full \methodshort{} pipeline; $N{=}5$ throughout. Each row adds one stage on top of the previous: random points replaces \methodshort{} with uniform interior sampling and skips labelling; \methodshort{} points restores geometry-aware selection but still skips labelling; $+$ VLM labels enables the verifier; $+$ filter enables confidence filtering at $\tau{=}0.7$. Bold marks the full pipeline (the headline run from Table~\ref{tab:main_tf}); \sbest{underline} marks the second-best.}
\label{tab:stage_ablation}
\begin{tabular}{@{}l l c@{}}
\toprule
\textbf{Stage} & \textbf{Description} & \textbf{cIoU} $\uparrow$ \\
\midrule
s1 & Bounding box only                                          & $69.2$ \\
s2 & $+$ Random interior points (no labels)                     & $65.4$ \\
s3 & $+$ \methodshort{} points (no labels)                      & $68.7$ \\
s4 & $+$ VLM labels (no filter)                                 & \sbest{$73.6$} \\
\textbf{s5} & \textbf{$+$ Confidence filter $\tau{=}0.7$ (full pipeline)} & $\boldsymbol{75.9}$ \\
\bottomrule
\end{tabular}
\end{table}

\paragraph{Where each stage helps.}
The stage walk reveals that the gain enters in clearly separable steps rather than smoothly. Replacing the box-only baseline with random interior points (s1$\rightarrow$s2) actually \emph{costs} $3.8$ cIoU: SAM treats every passed point as a positive prompt, so uninformative samples landing on co-occurring background or distractors pollute the support and shift the predicted mask away from the referent. Switching to geometry-aware \methodshort{} points (s2$\rightarrow$s3) recovers most of that loss ($+3.3$ cIoU), confirming that point selection alone---without semantic supervision---is enough to bring the prompt set back to the box-only floor by avoiding obviously bad positions. The biggest single jump is the VLM verifier (s3$\rightarrow$s4, $+4.9$ cIoU): converting a geometric set into per-point positive/negative labels gives SAM both anchors \emph{and} explicit exclusions, which is what disambiguates the referent from co-occurring objects inside the same box. Confidence filtering on top of labelling (s4$\rightarrow$s5, $+2.3$ cIoU) is the final stabiliser, suppressing the small fraction of low-confidence labels the verifier mislabels under grammar-guided decoding; Tables~\ref{tab:filter_tau} and~\ref{tab:filter_allowlist} dissect this stage further.

\begin{table}[h]
\centering
\small
\setlength{\tabcolsep}{6pt}
\caption{\textbf{Per-cue robustness on RefCOCO val.} Each row zeros one cue of the four-cue \methodshort{} selector under the diagnostic linear-penalty fusion that admits clean off-switches for $E$ and $H$ via $\lambda_E$ and $\lambda_H$, and for $G$ via $\sigma_G\!\to\!\infty$. We report $\Delta$cIoU relative to the all-cues-on baseline of this fusion; the absolute level under the diagnostic fusion is not directly comparable to the headline $C_{\text{mult}}$ runs in Tables~\ref{tab:main_tf} and~\ref{tab:ablations}.}
\label{tab:cue_ablation}
\begin{tabular}{@{}l l c@{}}
\toprule
\textbf{Setting} & \textbf{Cue removed} & $\Delta$\textbf{cIoU} vs.\ all-on \\
\midrule
all-on  & ---                                                  & $0.0$ \\
$\text{no}\_E$ & edge density ($\lambda_E{=}0$)                & \dneg{$-0.7$} \\
$\text{no}\_G$ & Gaussian prior ($\sigma_G{\to}\infty$)        & \dneg{$-1.5$} \\
$\text{no}\_H$ & local entropy ($\lambda_H{=}0$)               & \dpos{$+1.7$} \\
\bottomrule
\end{tabular}
\end{table}

\paragraph{Robustness to single-cue dropout.}
No single cue moves cIoU by more than $1.7$ points on RefCOCO val, supporting the robustness claim already made in §\ref{sec:experiments:ablations} (Hyperparameter sensitivity): \methodshort{} does not rely on any single cue. Removing the Gaussian prior costs the most ($-1.5$ cIoU), consistent with $G$'s role as a soft centroid prior that suppresses peripheral candidates close to the box border. Removing the edge-density penalty costs only $-0.7$ cIoU at unit weight. The small \emph{positive} $\Delta$ for $\text{no}\_H$ ($+1.7$) is a property of the unit-weighted linear-penalty diagnostic, not of the headline fusion: the entropy penalty at $\lambda_H{=}1$ is somewhat over-aggressive, which is consistent with the $C_{\text{msm}}$ sweep in §\ref{sec:experiments:ablations} preferring a different operating point rather than $H$ being unhelpful in general. The headline $C_{\text{mult}}$ uses $(1-E)$ and $(1-H)$ as smoothness/homogeneity factors with no scalar weights, so $E$ and $H$ contribute as fixed multiplicative penalties there and cannot be probed by zeroing a parameter; the diagnostic mode used in Table~\ref{tab:cue_ablation} exists precisely to make these per-cue interventions well-defined.

\paragraph{Filter sweep: confidence threshold.}
The verifier ablation in Table~\ref{tab:ablations} reports two threshold endpoints, $\tau{=}0$ and $\tau{=}0.7$. Table~\ref{tab:filter_tau} sweeps the interior, $\tau \in \{0.5, 0.6, 0.7, 0.8, 0.9\}$, on the multiplicative variant with the default allowlist $\mathcal{A}{=}\{\texttt{positive},\texttt{negative}\}$; we report cIoU on the full RefCOCO val and testB splits along with the fraction of labeled points discarded and the mean number of points passed to SAM ($N{=}5$ throughout). cIoU is non-monotone in $\tau$ on both splits and peaks at $\tau{=}0.8$; the $\tau{=}0.5$, $\tau{=}0.6$, $\tau{=}0.7$ rows sit within a $1.0$ cIoU band on val. This pattern is consistent with the calibration of grammar-guided Qwen2.5-VL: verbalized confidences below ${\sim}0.7$ are systematically mis-labeled and benefit from removal, but above ${\sim}0.7$ confidence and label correctness are essentially uncorrelated, so further filtering discards correct labels along with the wrong ones. Pushing past $\tau{=}0.8$ then begins to cost: at $\tau{=}0.9$, $\sim\!48\%$ of labeled points are discarded on val (mean kept prompts $2.51$, vs.\ $4.27$ at default), and a non-trivial sample fraction retains no prompts at all and falls back to SAM's box-only output ($69.2$ cIoU; Table~\ref{tab:ablations}). The both-labels cIoU at $\tau{=}0.9$ ends $-0.8$ below default on both splits, giving a clean illustration of the box-only-fallback tradeoff.

\begin{table}[h]
\centering
\small
\setlength{\tabcolsep}{6pt}
\caption{\textbf{Confidence-threshold sweep on RefCOCO val and testB.} cIoU at each $\tau$ on the default \methodshort\textsubscript{mult} configuration with $\mathcal{A}{=}\{\texttt{positive},\texttt{negative}\}$, $N{=}5$. ``\% filt.'' is the fraction of labeled points discarded; ``mean kept'' is the mean number of points passed to SAM. The \textbf{bold} row marks the paper default; \sbest{underlined} cells are the best within each split.}
\label{tab:filter_tau}
\begin{tabular}{@{}c c c c c c c@{}}
\toprule
$\tau$ & val cIoU & testB cIoU & val \% filt. & testB \% filt. & val mean kept & testB mean kept \\
\midrule
$0.5$ & $76.0$ & $69.0$ & $0.9$  & $1.0$  & $4.71$ & $4.75$ \\
$0.6$ & $75.2$ & \sbest{$73.0$} & $4.3$  & $3.1$  & $4.56$ & $4.57$ \\
$\boldsymbol{0.7}$ & $\boldsymbol{75.9}$ & $\boldsymbol{71.1}$ & $\boldsymbol{12.8}$ & $\boldsymbol{8.8}$ & $\boldsymbol{4.27}$ & $\boldsymbol{4.43}$ \\
$0.8$ & \sbest{$78.1$} & $71.9$ & $29.0$ & $20.6$ & $3.47$ & $3.67$ \\
$0.9$ & $75.1$ & $70.3$ & $47.8$ & $42.4$ & $2.51$ & $2.72$ \\
\bottomrule
\end{tabular}
\end{table}

\paragraph{Filter sweep: label allowlist.}
Table~\ref{tab:filter_allowlist} extends the sweep to the joint grid of $\tau \in \{0.7, 0.8, 0.9\}$ and $\mathcal{A} \in \{\{+,-\}, \{+\}, \{-\}\}$. The optimal allowlist is split-dependent. Val prefers positive-only at $\tau{=}0.8$ ($80.3$ cIoU, $+4.4$ over default), while testB prefers negative-only at the same $\tau$ ($72.1$ cIoU, $+1.0$); the two splits flip the sign of the allowlist effect, so committing to a single label class is brittle. SAM uses positives to \emph{anchor} and negatives to \emph{exclude}, and the relative importance of the two roles depends on the cluttered-box population of the split: testB has more distractors, so removing negatives hurts more there. Beyond $\tau{=}0.8$ all three allowlists move toward or past the box-only-fallback regime: positive-only retains only ${\sim}1.1$ prompts on val and ${\sim}1.2$ on testB at $\tau{=}0.9$, and the both-labels cell drops below default on both splits.

\begin{table}[h]
\centering
\small
\setlength{\tabcolsep}{6pt}
\caption{\textbf{Threshold $\times$ label-allowlist sweep on RefCOCO val and testB.} cIoU under three allowlists $\mathcal{A}$: both labels, positive-only ($\{+\}$), negative-only ($\{-\}$). $N{=}5$, multiplicative fusion, oracle off. The bold row is the paper default; \sbest{underlined} cells are the best within each split.}
\label{tab:filter_allowlist}
\begin{tabular}{@{}c c c c c c@{}}
\toprule
$\tau$ & $\mathcal{A}$ & val cIoU & testB cIoU & val mean kept & testB mean kept \\
\midrule
$\boldsymbol{0.7}$ & $\boldsymbol{\{+,-\}}$ & $\boldsymbol{75.9}$ & $\boldsymbol{71.1}$ & $\boldsymbol{4.27}$ & $\boldsymbol{4.43}$ \\
$0.7$ & $\{+\}$    & $76.3$ & $69.6$ & $1.46$ & $1.88$ \\
$0.7$ & $\{-\}$    & $73.1$ & $69.4$ & $2.70$ & $2.41$ \\
\midrule
$0.8$ & $\{+,-\}$  & $78.1$ & $71.9$ & $3.47$ & $3.67$ \\
$0.8$ & $\{+\}$    & \sbest{$80.3$} & $71.4$ & $1.33$ & $1.79$ \\
$0.8$ & $\{-\}$    & $77.2$ & \sbest{$72.1$} & $2.25$ & $2.05$ \\
\midrule
$0.9$ & $\{+,-\}$  & $75.1$ & $70.3$ & $2.51$ & $2.72$ \\
$0.9$ & $\{+\}$    & $79.0$ & $68.8$ & $1.09$ & $1.18$ \\
$0.9$ & $\{-\}$    & $77.4$ & $69.9$ & $1.51$ & $1.51$ \\
\bottomrule
\end{tabular}
\end{table}

\section{Per-stage runtime breakdown}
\label{app:runtime}

This appendix expands the brief efficiency discussion in Section~\ref{sec:experiments:efficiency}. We characterise \methodshort{}'s inference cost in terms of VLM calls per query, the dominant cost contributor in any training-free VLM\,+\,SAM pipeline and a hardware-independent measure of architectural efficiency, before reporting the per-stage wall-clock breakdown.

\paragraph{Why the per-query budget is exactly two VLM calls.} Under \methodshort{}'s default configuration (grammar-guided structured decoding in both the localization and point-labeling stages, no separate chain-of-thought reasoning pass, and batched per-image labeling), a single query results in (i)~one grammar-guided localization call producing the candidate set $\mathcal{B}$, and (ii)~one batched grammar-guided point-labeling call across all $(b_k, p)$ pairs. The two-call budget is therefore independent of the point budget $N$, the number of proposed boxes $K$, and the choice of fusion mode. The remaining stages of the pipeline (\methodshort{} point selection, point filtering, and SAM segmentation) are not VLM calls: cue computation and Soft-NMS run on the CPU, and SAM is a single feed-forward pass through the frozen segmenter.

\paragraph{Call-count comparison with related training-free pipelines.} Training-free referring-segmentation methods divide cleanly along their inference-time call budget. Decomposition pipelines such as RESAnything~\citep{resanything2025neurips} parse the referring expression into multiple semantic attributes and issue one VLM call per attribute. Loop-based pipelines such as SegLLM~\citep{SegLLM2024arxiv} and CoT-RVS~\citep{CoT-RVS2025arxiv} iteratively feed segmentation outputs back into the VLM for refinement, spending three or more VLM calls per refinement round. RL-trained single-shot methods such as Seg-Zero~\citep{SegZero2025arxiv} and SAM-R1~\citep{SAMR1_2025} issue a single VLM call at inference but require GRPO post-training on thousands of curated samples. \methodshort{} matches the inference-time call budget of the RL-trained category without any training, and issues fewer VLM calls than decomposition or loop-based methods on expressions of nontrivial length.

\paragraph{Per-stage wall-clock.} Table~\ref{tab:runtime} reports the per-query wall-clock decomposition of \methodshort{} on RefCOCO val with Qwen2.5-VL-7B + SAM, measured on a single A6000. Means and standard deviations are computed across the val split. Total cost is dominated by the two VLM calls (localization and point labeling, together ${\sim}95\%$ of wall-clock); \methodshort{}'s geometry-aware point selection contributes only $1.3\%$ and confidence filtering is below $0.1$\,ms.

\begin{table}[h]
\centering
\small
\caption{\textbf{Per-query wall-clock breakdown on RefCOCO val (single A6000).} Confidence filtering was disabled in this measurement; its arithmetic cost is below $0.1$\,ms.}
\label{tab:runtime}
\begin{tabular}{@{}lrrr@{}}
\toprule
\textbf{Stage} & \textbf{Mean (ms)} & \textbf{Std (ms)} & \textbf{\% of total} \\
\midrule
Localization (VLM)                        & $3290.1$        & $952.9$         & $66.5\%$ \\
\textbf{Point selection (\methodshort{})} & $\mathbf{62.2}$ & $\mathbf{14.3}$ & $\mathbf{1.3\%}$ \\
Point labeling (VLM)                      & $1413.7$        & $196.9$         & $28.6\%$ \\
Point filtering                           & $<0.1$          & $<0.1$          & $0.0\%$ \\
SAM segmentation                          & $182.5$         & $23.1$          & $3.7\%$ \\
\midrule
\textbf{Total}                            & $\mathbf{4948.5}$ &              & $\mathbf{100.0\%}$ \\
\bottomrule
\end{tabular}
\end{table}

The ratio of \methodshort{} point-selection cost to total VLM cost is $62.2\,/\,(3290.1+1413.7) \approx 1.3\%$, confirming that the per-query budget is dominated by the two structured-decoding VLM calls and that \methodshort{}'s overhead is negligible despite being more expensive than uniform random sampling inside the box.


\end{document}